\documentclass[Afour,sageh,times]{sagej}

\usepackage{moreverb,url}
\usepackage[lined,boxed]{algorithm2e}
\usepackage{amsmath} 
\usepackage{mathtools}
\usepackage{xcolor}
\usepackage{amssymb}  
\usepackage{subcaption}
\usepackage{textcomp}
\usepackage{booktabs}
\usepackage{flushend} 
\usepackage{balance}
\usepackage{multirow}
\usepackage{makecell}
\usepackage[colorlinks,bookmarksopen,bookmarksnumbered,citecolor=blue]{hyperref}

\newcommand\BibTeX{{\rmfamily B\kern-.05em \textsc{i\kern-.025em b}\kern-.08em
T\kern-.1667em\lower.7ex\hbox{E}\kern-.125emX}}

\def\volumeyear{2025}

\begin{document}

\runninghead{BIM-Loc}

\title{BIM-Loc: BIM-Integrated Discrepancy-Aware LiDAR-based Indoor Localization}

\author{Yinqiang Zhang\affilnum{1}, Liang Lu\affilnum{1}, Yipeng Pan\affilnum{1}, Maolin Lei\affilnum{3}, Yuhan Xie\affilnum{1}, Zhanteng Xie\affilnum{1}, Xiaowei Luo\affilnum{2}, Jia Pan\affilnum{1}}

\affiliation{\affilnum{1}Department of AI \& Data Science, University of Hong Kong (HKU), Hong Kong, China. \\
\affilnum{2}Department of Architecture and Civil Engineering, City University of Hong Kong (CityU), Hong Kong, China. \\
\affilnum{3}Humanoids and Human Centered Mechatronics Research Line, Italian Institute of Technology (IIT) Genoa, Italy.  
}

\corrauth{Jia Pan, the University of Hong Kong.}
\email{jpan@cs.hku.hk}

\begin{abstract}
Accurate and robust localization is a fundamental requirement for service and inspection robots, particularly in feature-sparse indoor environments where traditional systems struggle due to a lack of distinct landmarks. While prior maps can enhance robustness, precise and compact maps capturing real-world details are often unavailable for new or frequently changing environments. This paper presents BIM-Loc, a novel discrepancy-aware LiDAR-based localization method that directly integrates Building Information Models (BIM) from the design phase. BIM-Loc simultaneously estimates trajectories aligned with the BIM coordinate system and identifies discrepancies between real-world observations and the as-designed BIM in an online fashion. Our core contributions include: (1) a novel multi-hit ray casting strategy for efficient BIM-point data association and projection of 3D observations into 2D texture space; (2) a pose graph optimization framework with BIM-integrated factors that enforces consistency among odometry, sequential scans, and BIM structures; and (3) a hierarchical Bayesian inference module that incrementally updates a continuous 2D surface representation for discrepancy detection, propagating updates from the pixel to the structure level. Extensive evaluations in both simulation and real-world applications demonstrate that BIM-Loc significantly outperforms state-of-the-art map-based methods in localization accuracy and robustness. More experimental results are available at our project website: \url{https://bim-loc.github.io/bim-loc}.
\end{abstract}


\maketitle 

\section{Introduction}
\label{sec:intro}
\begin{figure}[t]
    \centering
    \includegraphics[width=1.0\linewidth]{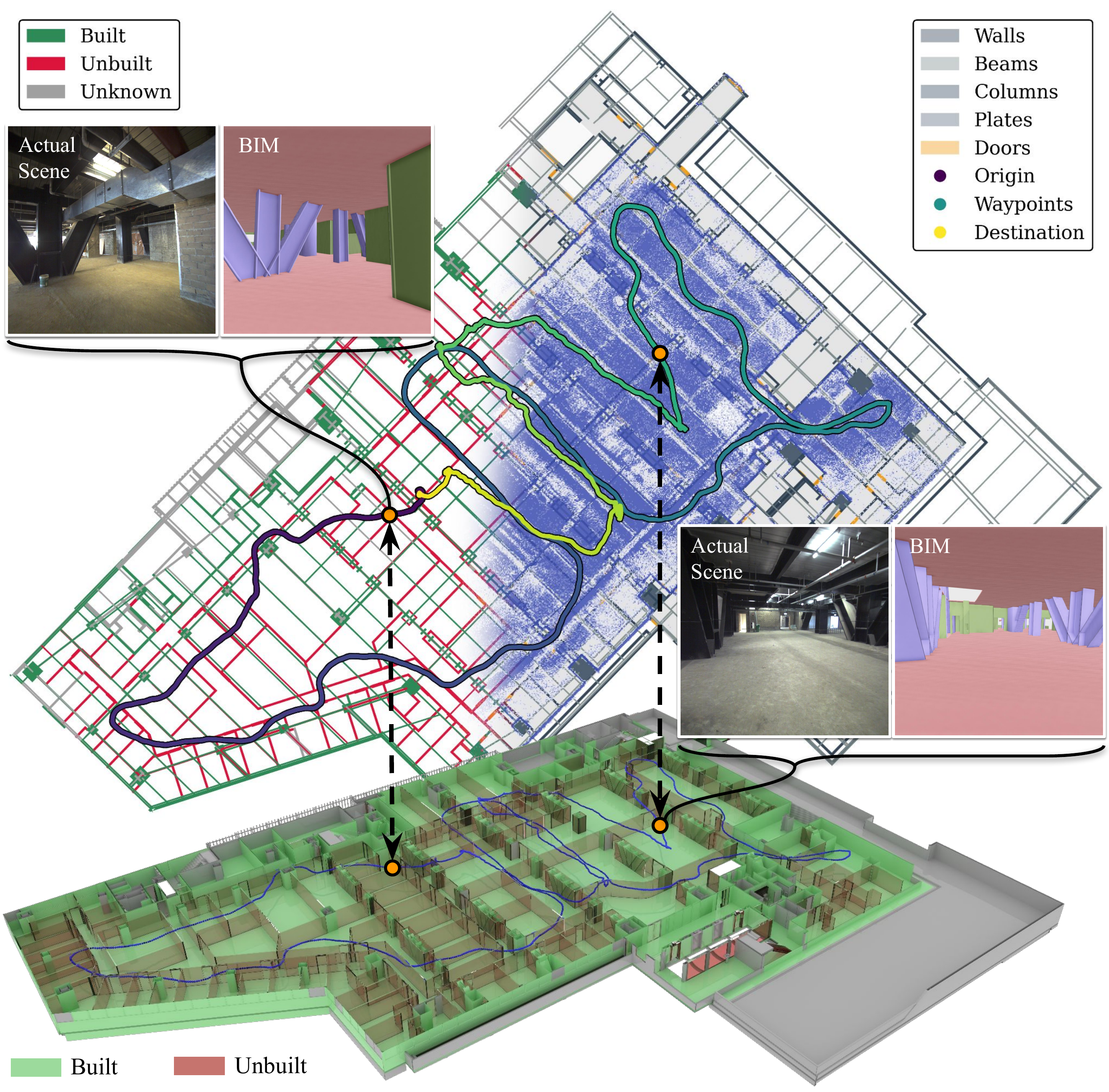}
    \caption{
    Visualization of BIM-Loc's outputs. The right panel displays blue reconstructed points aligned with a projected 2D BIM model. The left panel and bottom 3D meshes show the discrepancy map generated from localization. Two snapshots demonstrate localization performance by comparing actual scenes with corresponding BIM rendering.
    }
    \label{fig:hero-image}
\end{figure}

Accurate localization is crucial for applications in extreme environment exploration, remote sensing, and service robotics~\citep{Cadena2016TRO, Ebadi2022ArXiv, Yarovoi2024AIC, Chen2025RAL}. For robots acting as service providers or inspectors in indoor settings, robust localization is essential throughout a building's life cycle, from construction sites to completed structures. However, ensuring stable localization for inspection and service tasks presents significant challenges. Indoor environments often exhibit repetitive, human-designed geometric configurations that lack sufficient distinctive features for robust localization under varying conditions. Furthermore, unlike outdoor scenarios where GNSS provides a stable global reference~\citep{Wang2022RAL}, GNSS-denied indoor environments lack equivalent signals to correct the accumulated drift errors inherent in long-term navigation.

To enhance indoor localization robustness, one common approach uses global place descriptors~\citep{Yin2024IJCV} to construct loop closures. Coupled with pose-graph optimization~\citep{Grisetti2010ITSM, Kenji2024RAS} or bundle adjustment~\citep{Liu2023RAL, Liu2023TRO, Li2024bIROS}, these methods can reduce drift within closed loops. However, they require active loop-closure detection and are ineffective in one-way scenarios like long corridors or tunnels. Another line of research leverages map priors, with the specific approach varying by sensor modality. For example, radio-frequency-based sensors (e.g., WIFI, BLE, RFID) can create reference sensor network maps~\citep{Roohi2025CN}. While robust, these methods require pre-deployed anchor sensors and typically offer insufficient accuracy for precise robotic inspection in dynamic environments.

Topological and structural metric maps have also been extensively investigated. Recent studies utilize point-cloud-based maps~\citep{Koide2024ICRA, Hu2024TMech} to bound localization errors. However, raw point clouds are not scalable for large buildings. Within the Architecture, Engineering, and Construction (AEC) industry, 2D floor plans~\citep{Gao2022ICRA, Xie2023RAL, Li2024IROS} and 3D BIM models~\citep{Torres2023ECPPM} are used throughout the building life cycle. Unlike raw point clouds, the high-level semantic and geometric abstractions in BIM and floor plans significantly reduce memory consumption, making them ideal for edge computing devices. While 2D floor plan localization requires additional height estimation, BIM's comprehensive 3D descriptions naturally support 3D localization in multi-floor buildings. A critical challenge, however, is that BIM models are typically ``as-designed'' and do not reflect the actual, often divergent, state of construction sites or renovated buildings. This BIM-reality discrepancy severely hinders BIM integration for localization, a problem that remains an open research question explored by only few researchers.

Driven by these challenges and the advantages of BIM, this paper proposes BIM-Loc, a BIM-integrated, discrepancy-aware method for robust 3D LiDAR-based indoor localization. BIM-Loc effectively bounds accumulated drift even under significant discrepancies between the BIM model and the real world. Instead of relying on point-based maps, it uses only the as-designed BIM. We reformulate the map-discrepancy-affected localization problem into two coupled subproblems: \textit{BIM-aided trajectory optimization} and \textit{hierarchical discrepancy detection}. A novel \textit{multi-hit ray casting strategy} enables efficient ray-BIM association for deep BIM integration. By iteratively solving these subproblems, BIM-Loc simultaneously estimates a BIM-aligned trajectory and identifies the current BIM status online. The visualization of BIM-Loc's outputs are illustrated in Figure~\ref{fig:hero-image}. Our core contributions are:
\begin{itemize}
    \item A novel multi-hit ray-casting module that establishes data association for both local surface sampling and texture-space mapping, underpinning trajectory optimization and discrepancy detection.
    \item A BIM-aided trajectory optimization module that ensures consistency between the BIM and real-world coordinate systems through a pose graph framework with custom factors.
    \item An online BIM status update module for discrepancy detection. Using sparse scan data, it employs Bayesian kernelized inference to incrementally update a 2D texture-space representation of discrepancies, which are then propagated to the structure level via a hierarchical Bayesian network.
    \item Extensive validation on a custom BIM-robot simulation platform and real-world datasets from both construction sites and building interiors, demonstrating superior performance over state-of-the-art methods.
\end{itemize}

The remainder of this paper is structured as follows: Section~\ref{sec:related-work} reviews related work on SLAM optimization, map-aided localization, and AEC applications. Section~\ref{sec:method} details the BIM-Loc framework. Section~\ref{sec:experiments} presents experimental results. Section~\ref{sec:discussion} discusses the design and advantages of our approach, and Section~\ref{sec:futurework} outlines limitations and future work.

\section{Related Work}
\label{sec:related-work}
\subsection{LiDAR Localization and Mapping}
\label{sec:slam-work}
LiDAR-based SLAM provides centimeter-level localization for autonomous vehicles and UAVs, a capability critical in GNSS-denied indoor environments. Typically, LiDAR SLAM systems comprise a front-end for odometry (via scan matching) and a back-end for map optimization that enforces consistency among scans~\citep{Cadena2016TRO, Ebadi2022ArXiv}.

For robust scan matching, various geometric features are used within the Iterative Closest Point (ICP) paradigm. For example, KISS-ICP~\citep{Vizzo2023RAL} uses point-to-point ICP with adaptive thresholds, while other works employ point-to-plane metrics~\citep{Ferrari2024RAL} or combine them via covariance matrices~\citep{Kenji2021ICRA}. Feature-based methods like LOAM~\citep{Zhang2014RSS} and Lego-LOAM~\citep{Shan2018IROS} extract corner and planar features for improved performance. FAST-LIO2~\citep{Xu2022TRO} enables direct plane matching using an iKd-Tree for efficient search. NV-LIOM~\citep{Chung2024RAL} incorporates surface normal vectors for better indoor performance. BIM-Loc is agnostic to the front-end odometry, allowing it to seamlessly integrate with various algorithms to bound their drift.

For consistency-oriented map optimization, Pose-Graph Optimization (PGO) and Bundle Adjustment (BA) are standard techniques. Factor graphs are widely used in PGO; for instance, LIO-SAM~\citep{Shan2020IROS} formulates LiDAR-inertial odometry within a factor graph using GTSAM~\citep{gtsam}. For multi-session SLAM, LAT-OM~\citep{Zou2024JFR} uses STD~\citep{Yuan2023ICRA} loop detection for long-term mapping. In contrast, BA achieves finer feature-level consistency. Works like~\cite{Zhou2021RAL, Liu2023TRO} proposed global plane adjustment to refine plane parameters and keyframe poses jointly, later extended to lines, cylinders, and curves~\citep{Zhou2022RAL, Li2024bIROS}. The hierarchical framework HBA~\citep{Liu2023RAL} balances real-time performance and precision in large-scale tasks by combining the idea of PGO and BA. Following this paradigm, BIM-Loc designs novel graph factors to tightly integrate as-designed BIM models into a PGO framework, enforcing both inter-scan and scan-BIM consistency constraints from the front-end odometry.

\begin{figure*}[t]
    \centering
    \includegraphics[width=1.0\linewidth]{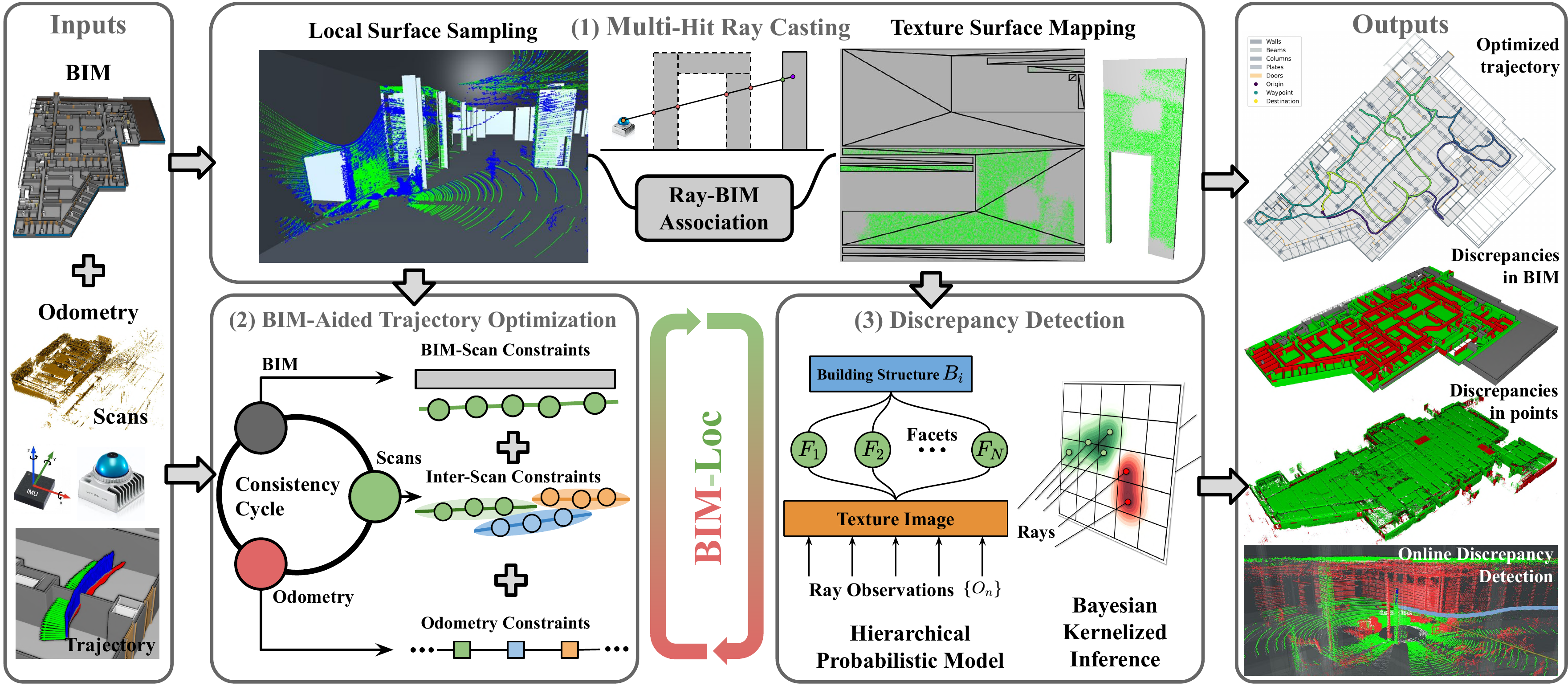}
    \caption{
Framework of the proposed BIM-Loc method. The system processes as-designed BIM models and front-end odometry measurements through three core modules: (1) multi-hit ray casting for BIM-observation association, (2) BIM-aided trajectory optimization, and (3) BIM-observation discrepancy detection. Using an iterative strategy, the method outputs an optimized trajectory and discrepancy maps, relying solely on design-phase BIM models.
    }
    \label{fig:overall-framework}
\end{figure*}

\subsection{Indoor Localization with Map Priors}
\label{sec:prior-work}
Prior maps serve as a reliable reference in localization. While high-definition maps with GNSS are well-established in autonomous driving, various map types are used indoors due to the absence of satellite signals. Schematic floor plans are a common choice for indoor localization. Wang et al.~\citep{Wang2019IROS} proposed a factor graph-based global localization method that utilizes wall intersections as features, incorporating robust data association through pairwise measurement consistency checks and a max-mixtures error model. FP-Loc~\citep{Gao2022ICRA} uses a custom nearest-neighbor search for efficient wall and column lookup from floor plans, while AG-Loc~\citep{Xie2023RAL} leverages area graphs extracted from floor plans for lifelong localization given a broad initial pose estimate. Compared to schematic floor plans, prior maps reconstructed from recorded point clouds provide more direct geometric information for map-based localization. 2D grid maps have been utilized for global localization and trajectory refinement~\citep{Hess2016ICRA}. More recently, research has increasingly focused on pre-built 3D point cloud maps that offer rich geometric constraints for 6-DoF localization. For instance, Koide et al.~\citep{Koide2024ICRA} use reconstructed point clouds as reference maps for a range-inertial localization with a sliding-window factor graph. PALoc~\citep{Hu2024TMech} utilizes point-cloud-based prior maps to generate ground-truth-level trajectories with uncertainty estimation in degraded indoor environments. Other works use representations like normal distributions~\citep{Einhorn2015RAS} or meshes~\citep{Dreher2021ISARC, Mock2024IROS} to reduce memory footprint of the prior maps.

In the AEC industry, several works have been proposed to integrate BIM into localization. Torres et al.~\citep{Torres2023ECPPM} project BIM into 2D occupancy grid maps for particle-filter-based robust localization. An extension, BIM-SLAM~\citep{Vega2024CR}, supports multi-session tasks by exploiting loop detection, pose-graph optimization, and change detection. {LIO-BIM~\citep{Stuhrenberg2025AEI} couples LiDAR-inertial odometry with BIM through dual scan-to-BIM matching. The BIM/IFC files are processed to generate plain and corner features from point clouds.} Another approach converts BIM into semantic-enhanced point clouds to keep both geometric and semantic information so that a coarse-to-fine semantic localization can align laser points to the map via ICP~\citep{Yin2023AIC}. While these works demonstrate feasibility of BIM-aided SLAM, they do not fully and directly integrate BIM into the SLAM framework. Instead, they rely on points sampled from BIM surfaces, which incurs significant memory overhead that hinders deployment on resource-constrained devices. More importantly, these methods largely overlook the substantial BIM-reality discrepancies prevalent in construction and renovation scenarios, limiting their practical applicability in real-world settings where design models diverge from as-built conditions. {Recent efforts address such discrepancies through either continuous deviation estimation~\citep{Shaheer2025RAL}, which excels in structured environments via global graph matching and hierarchical semantic scene graphs, or probabilistic dynamic object removal~\citep{Jia2025TMM}.}

\subsection{Discrepancy-Aware Mapping}
Identifying and managing discrepancies in sensor measurements across multiple sessions has been widely investigated. While several studies provide multi-modal datasets for change detection~\citep{Park2021IROS, Lambert2021NIPS}, applying these techniques to robot localization remains challenging. The primary sources of discrepancies include dynamic objects and environmental changes. To distinguish dynamic objects from static backgrounds, a popular solution is employing set difference operations through map-to-map complement extraction~\citep{Wellhausen2017SSRR}. Visibility-based approaches prove more robust against localization errors. For example, Removert~\citep{Kim2020IROS} uses multi-scale range images to reconstruct static maps. Extending this to multi-session mapping, LT-mapper~\citep{Kim2022ICRA} provides a modular framework for dynamic change detection and multi-layer map management without requiring perfect initial alignment. For highly dynamic environments, DynamicFilter~\citep{Fan2022ICRA} combines visibility-based and map-based approaches for dynamic object removal, with subsequent extensions enabling long-range open-world navigation in fully dynamic settings~\citep{Fan2025JFR}. 

Recent research has also addressed map comparison challenges. Holoch et al.~\citep{Holoch2022IROS} proposed creating, comparing, and merging new maps upon re-localization to handle lifelong localization failures in dynamic environments, by detecting invalid merges caused by perceptual aliasing. Rozsa et al.~\citep{Rozsa2020Springer} introduced a feature-based, comparison-free change detection method for SLAM maps that uses ICP-like frame matching and pose graph optimization to identify alterations through residual analysis. For dynamic environment modeling, Nobre et al.~\citep{Nobre2018ICRA} enhanced traditional sparse maps by incorporating feature persistence and spatio-temporal correlations. This probabilistic framework captures relationships between nearby features to compute joint posteriors over feature persistence, improving online data association for localization. 

{Recent work on self-updating HD maps demonstrates the limitations of binary discrepancy detection, modeling changes instead as atomic operations, such as geometry edits (e.g., boundary vertex shifts) and marking edits, rather than simple existence checks~\citep{Wild2025ICCV}.} In the AEC industry, discrepancy checking is essential for construction progress monitoring~\citep{Meyer2022AIC, Huang2022AIC, ZHANG2025AIC}. Existing methods for verifying discrepancies between BIM models and real scenes typically rely on dense 3D point clouds and spatial voxelization. For instance, Meyer et al.~\citep{Meyer2022AIC} model laser range impacts on spatial occupancy using belief functions, while Huang et al.~\citep{Huang2022AIC} extend this framework by integrating semantic information with UAV-acquired photogrammetric point clouds.
In contrast, our BIM-Loc method introduces a novel ray-casting strategy that performs discrepancy detection directly in surface texture space, which significantly reduces memory overhead compared to point cloud or 3D voxel representations.

\section{Methodology}
\label{sec:method}
\subsection{Problem Formulation}
\label{sec:problem-formulation}
For a sensor suite containing a LiDAR sensor and an Inertial Measurement Unit (IMU), we define a sensor frame whose pose is iteratively estimated by front-end odometry. The sensor's pose at time step $k$ is denoted as $\boldsymbol{x}_k \coloneqq[\mathbf{R}_k|\boldsymbol{t}_k]$, where  $\boldsymbol{t}_k \in \mathbb{R}^3$ is the translation and $\mathbf{R}_k\in\text{SO}(3)$ is the orientation. The trajectory over $K$ time steps is $\mathcal{X} \coloneqq \{\boldsymbol{x}_1, \cdots, \boldsymbol{x}_K\}$. Each observation at step $k$, denoted $\boldsymbol{z}_k \coloneqq \{\tilde{\boldsymbol{x}}_k, \mathcal{P}_k\} \in \mathcal{Z}$, consists of the estimated sensor pose $\tilde{\boldsymbol{x}}_k$ and the corresponding scan points $\mathcal{P}_k$.

As map priors, we use as-designed Building Information Modeling (BIM) models $\mathcal{B}$, which contain structural components $\{\boldsymbol{b}_n \mid \boldsymbol{b}_n \in \mathcal{B}, n=1,\cdots,N\}$ like walls, columns, beams, and coverings. To account for discrepancies between the BIM model and reality, we introduce an indicator set $\mathcal{L}$. Each element  $\mathbb{I}_n \in \{0, 1\}$ in this set reflects the consistency of a BIM structure  $\boldsymbol{b}_n$ with its sensor data. The subset of BIM structures that exist in reality can then be inferred as $\mathcal{B}_\text{real} \coloneqq \{\boldsymbol{b}_n \mid \mathbb{I}_n = 0\}$.

This paper tackles the challenge of localization when map discrepancies exist. Our approach simultaneously estimates the motion trajectory $\hat{\mathcal{X}}$ and the BIM-reality discrepancies $\hat{\mathcal{L}}$ from noisy sensor measurements $\mathcal{Z}$ and the BIM model $\mathcal{B}$ in an online fashion. We formulate this as a Maximum a Posteriori (MAP) estimation problem:
\begin{equation}
    \begin{split}
        (\hat{\mathcal{X}}, \hat{\mathcal{L}}) & = \arg\max_{\mathcal{X}, \mathcal{L}}\Big[p(\mathcal{X}, \mathcal{L} \mid \mathcal{Z}, \mathcal{B})\Big] \\ 
                                            & = \arg\max_{\mathcal{X}, \mathcal{L}}\Big[p(\mathcal{X})\cdot p(\mathcal{Z}\mid \mathcal{X}, \mathcal{L}, \mathcal{B})\cdot p(\mathcal{L} \mid \mathcal{B})\Big] \ .
    \end{split}
    \label{eq:map-framework}
\end{equation}
Here, $p(\mathcal{X})$ is the trajectory prior from the front-end odometry, defined as $p(\mathcal{X})=p(\boldsymbol{x}_0)\prod_k p(\boldsymbol{x}_{k} \mid \boldsymbol{x}_{k-1})$, where $p(\boldsymbol{x}_0)$ is the initial pose prior. The term $p(\mathcal{L} \mid \mathcal{B})$ is the prior for the discrepancy indicators, which can incorporate external knowledge. For example, structures confirmed as complete by human inspection can be assigned a low initial probability of discrepancy. The likelihood $p(\mathcal{Z}\mid \mathcal{X}, \mathcal{L}, \mathcal{B})$ is decomposed sequentially as:
\begin{equation}
    p(\mathcal{Z}\mid\mathcal{X}, \mathcal{L}, \mathcal{B}) = \prod_{n=1}^N\prod_{k=1}^K \ p(\boldsymbol{z}_k\mid \boldsymbol{x}_k, \boldsymbol{b}_n)^{(1-\mathbb{I}_n)} \ , 
    \label{eq:consistency-constraints}
\end{equation}
where $p(\boldsymbol{z}_k\mid \boldsymbol{x}_k, \boldsymbol{b}_n)$ is the likelihood for a single observation, and the discrepancy indicator $\mathbb{I}_n$ controls its contribution to the overall objective.

Maximizing this likelihood is equivalent to minimizing the residual error $\mathcal{E} = \mathcal{E}_\text{odom}+\mathcal{E}_\text{consistency}$, which balances the odometry prior $\mathcal{P}(\mathcal{X})$ against the BIM-consistency $p(\mathcal{Z} \mid \mathcal{X}, \mathcal{B}, \mathcal{L})$. The two residual components are:
\begin{equation}
    \begin{split}
        \mathcal{E}_\text{odom} &=r(\boldsymbol{x}_0)+\sum_{k=1}^K r(\boldsymbol{x}_k\mid \boldsymbol{x}_{k-1}) \ , \\
        \mathcal{E}_\text{consistency}&=\sum^K_{k=1}\sum_{n=1}^N\underbrace{\delta(n, k)}_{(1)} \cdot \underbrace{r(\mathcal{P}_k, \boldsymbol{x}_k, \boldsymbol{b}_n)}_{(2)}\cdot\underbrace{(1-\mathbb{I}_n)}_{(3)} \ , 
    \end{split}
    \label{eq:problem-costs}
\end{equation}
Here, the term $\mathcal{E}_\text{odom}$ includes residuals from the initial pose $r(\boldsymbol{x}_0)$ and sequential odometry estimates $r(\boldsymbol{x}_k \mid \boldsymbol{x}_{k-1})$. 
The term $\mathcal{E}_\text{consistency}$ quantifies the alignment between observation $\boldsymbol{z}_k$ and BIM structure $\boldsymbol{b}_n$ given pose $\boldsymbol{x}_k$, with $r(\mathcal{P}_k, \boldsymbol{x}_k, \boldsymbol{b}_n)$ measuring the consistency error. 
The association indicator $\delta(n, k)$ is $1$ if scan $\mathcal{P}_k$ overlaps with structure $\boldsymbol{b}_n$, and the term \((1-\mathbb{I}_n)\) activates the residual only for structures believed to be consistent with reality ($\mathbb{I}_n=0$).

As shown in Figure~\ref{fig:overall-framework}, the BIM-Loc framework computes the three components of $\mathcal{E}_\text{consistency}$ through dedicated modules: (1) the association indicator \(\delta(\cdot)\) is determined by a novel multi-hit ray casting strategy (Section~\ref{sec:multi-hit}); (2) consistency residuals $r(\mathcal{P}_k, \boldsymbol{x}_k, \boldsymbol{b}_n)$ are optimized within a BIM-aided trajectory module (Section~\ref{sec:bim-factor}) to estimate $\hat{\mathcal{X}}$. and (3) the discrepancy indicator $\mathbb{I}_n$ is updated by a discrepancy detection module (Section~\ref{sec:discrepancy-identification}) to estimate $\hat{\mathcal{L}}$. The online localization is solved by iterating between trajectory estimation and discrepancy detection. The following sections detail each module. A summary of notations used in the algorithm details is provided in Table~\ref{tab:symbol-description}.

\RestyleAlgo{ruled}
\begin{algorithm}[t]
    \caption{Multi-hit ray casting}\label{algo:multi-hit} 
    \SetAlgoLined
    \SetKwFunction{FMain}{MainFunction}
    \SetKwFunction{FRayCast}{RayCasting}
    \SetKwFunction{FStatus}{StatusEvaluation}
    \SetKwFunction{FUpdate}{RayUpdate}
    \KwIn{LiDAR rays $\mathcal{R}$, BIM models $\mathcal{B}$}
    \KwOut{Hit set $\mathcal{H}$, Miss set $\mathcal{M}$}
    $\mathcal{H} \gets \{\}, \ \mathcal{M} \gets \{\}$\;
    \While{$|\mathcal{R}|>0$}{
        $\mathcal{R}^{\prime} \gets \{ \}$\;
        \ForEach{$r \in \mathcal{R}$}{
            $(u, v,\boldsymbol{p^{\prime}},\boldsymbol{n}) \gets $\FRayCast{$r, \mathcal{B}$}\;\label{func:ray-casting}
            $\sigma \gets $\FStatus{$u,v,\boldsymbol{p^{\prime}},\boldsymbol{n}$}\;\label{func:status-evaluation}
            \uIf{$\sigma=1$}{
                $\mathcal{H} \gets \mathcal{H} \cup \{(u, v,\boldsymbol{p^{\prime}},\boldsymbol{n})\}$\;
            }
            \uElseIf{$\sigma=0$}{
                $\mathcal{M} \gets \mathcal{M} \cup \{(u, v,\boldsymbol{p^{\prime}},\boldsymbol{n})\}$\;
                $r \gets $\FUpdate{$r$, $\boldsymbol{p^{\prime}}$}\;\label{func:ray-update}
                $\mathcal{R}^{\prime} \gets \mathcal{R}^{\prime} \cup \{r\}$\;
            }
        }
        $\mathcal{R} \gets \mathcal{R}^{\prime}$\;
    }
\end{algorithm}

\subsection{Multi-Hit Ray Casting}
\label{sec:multi-hit}
\begin{figure}[t]
    \centering
    \includegraphics[width=1.0\linewidth]{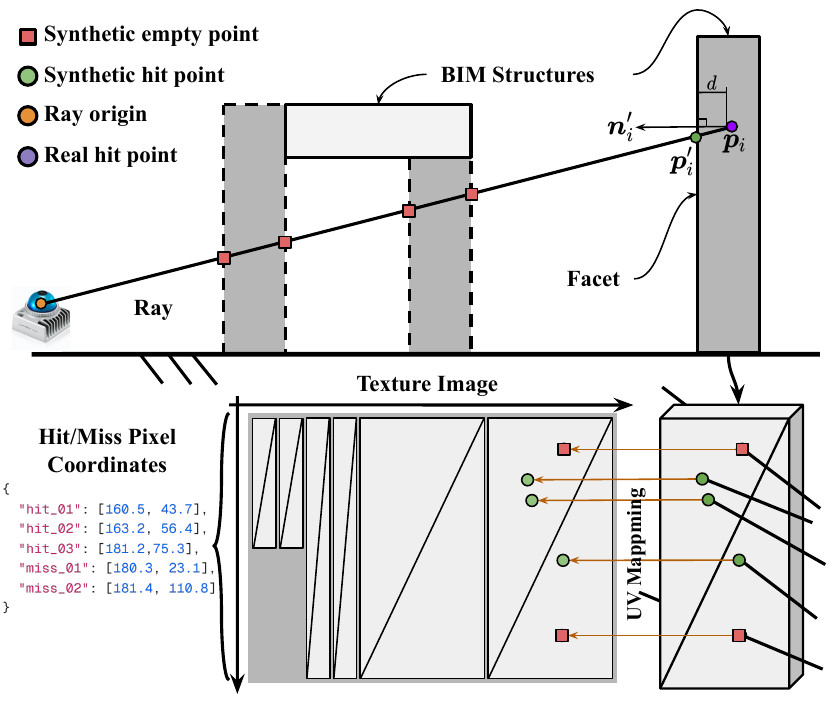}
    \caption{
    Multi-hit ray casting in BIM coordinate frame. The process transforms real-world point clouds into the BIM frame and performs ray casting until reaching each point's measured position. This provides local surface samples and texture mappings to BIM-Loc.
    }
    \label{fig:multi-hit}
\end{figure}

The paths of casted rays reflect real-world surface conditions, making them ideal for associating sensor observations with BIM models. Building on this principle, we propose a novel multi-hit ray casting model, illustrated in Figure~\ref{fig:multi-hit}.

For a given point cloud (omitting time subscript $k$), each point is treated as a ray defined by its origin, measured distance, and unit direction vector. We first transform these rays into the BIM coordinate system. A ray casting operation then simulates each ray's path through the BIM model, starting from the sensor origin and terminating {slightly behind the real-world measured endpoint with a extended length to deal with sensor noise. Under this setting, even the ray would normally stop slightly before the BIM surface, the designed extended length forces an intersection.}

Unlike standard methods that stop at the first intersection, our strategy evaluates multiple intersections along the ray's path until it reaches the neighborhood of the endpoint. This process is detailed in Algorithm~\ref{algo:multi-hit}. For each intersection with a BIM surface (facet), the \texttt{StatusEvaluation} function computes the point-to-surface distance $d$ (see Figure~\ref{fig:multi-hit}, upper right). If the ray has not yet reached its endpoint neighborhood, the \texttt{RayUpdate} function sets the intersection point as a new origin and continues casting. This iterative process continues until all rays are processed, generating a set of synthetic counterpart points in the BIM coordinate system.

For each intersection, the \texttt{RayCasting} function returns the following information: (1) association: the indices of the struck BIM structure and its specific planar facet; (2) synthetic surface point: a hit point $\boldsymbol{p}^\prime$ and its normal vector  $\boldsymbol{n}^\prime$ on the BIM surface; (3) projection data: barycentric coordinates $(u, v)$ and a binary intersection status $\sigma$. The barycentric coordinates result from projecting the 3D intersection onto a 2D texture space using a pre-computed UV-mapping dictionary (generated by the xatlas library~\citep{Young2025XATLAS}). This projection strategy avoids storing numerous 3D points, significantly reducing memory consumption. The intersection status $\sigma$ is 1 (hit) if the distance from the synthetic point to the real surface is below a threshold, and 0 (miss) otherwise. The collected data is then organized into a hit set $\mathcal{H}$ and a miss set $\mathcal{M}$ based on this status.

In summary, the multi-hit ray casting module contributes to the BIM-Loc framework in three critical ways. {First, it establishes a efficient and accurate association between BIM structures and point clouds, which is comparable with traditional nearest-neighbor search.} Second, the synthetic hit points with normals in $\mathcal{H}$ serve as local samples of BIM surfaces, which are crucial for the BIM-aligned trajectory estimation in Section~\ref{sec:bim-factor}. Finally, the barycentric coordinates enable the projection of 3D observations into a 2D texture space, which is fundamental to the discrepancy detection method in Section~\ref{sec:discrepancy-identification}.

\begin{figure*}[t]
    \centering
    \includegraphics[width=1.0\linewidth]{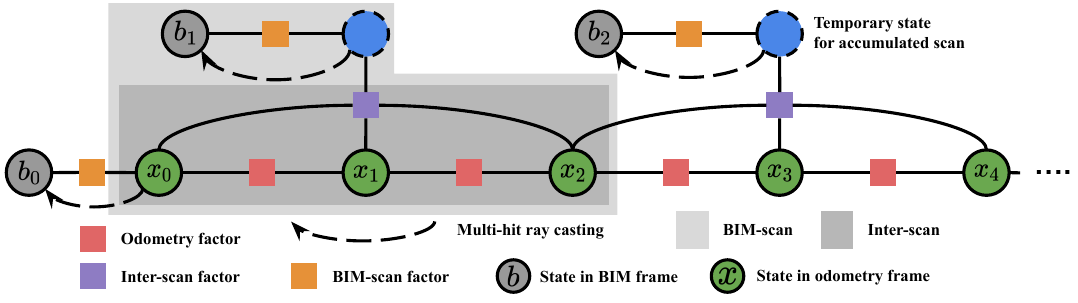}
    \caption{
    Schema of the factor graph for trajectory optimization. Scans are transformed into a common temporary state, then processed via batch multi-hit ray casting. Synthetic points are represented in the BIM coordinate system. The pose graph incorporates three types of constraint factors to balance inter-scan consistency and BIM-scan alignment during online optimization.
    }
    \label{fig:factor-graph}
\end{figure*}

\subsection{BIM-Aided Trajectory Optimization}
\label{sec:bim-factor}
This section describes the trajectory optimization that maintains consistency between sensor observations and the BIM model. We formulate the residuals from Eq.~\ref{eq:problem-costs} as a pose factor graph. This graph combines two components: (1) motion priors from front-end odometry  $\mathcal{E}_\text{odom}$, derived from $p(\mathcal{X})$ and (2) BIM-observation consistency  $\mathcal{E}_\text{consistency}$, derived from the likelihood $p(\mathcal{Z}\mid\mathcal{X}, \mathcal{B},\mathcal{L})$ in Eq.~\ref{eq:consistency-constraints}.

For the trajectory optimization sub-problem, the association indicator $\delta(k, n)$ is provided by the multi-hit ray casting module (Section~\ref{sec:multi-hit}), while the discrepancy indicator  $\mathbb{I}_{n}$  is held fixed. This allows us to isolate the effective composite residual for independent analysis:
$\sum_{\delta(k,n)=1, \mathbb{I}_n=0}r(\mathcal{P}_k, \boldsymbol{x}_k, \boldsymbol{b}_n) \coloneqq r(\mathcal{Z}, \mathcal{B})$.

\begin{figure}[t]
    \centering
    \includegraphics[width=1.0\linewidth]{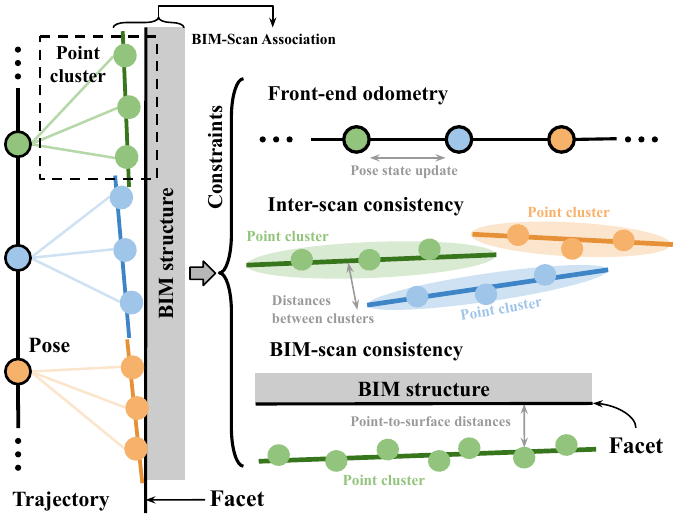}
    \caption{
Definition of constraints for BIM-aided trajectory optimization. Multi-hit ray casting associates scans with BIM structures. The optimization incorporates three constraints: front-end odometry, inter-scan consistency, and scan-BIM alignment, which jointly determine the trajectory estimation.
    }
    \label{fig:point-cluster}
\end{figure}

The multi-hit ray casting operation establishes data association by assigning each point in the hit set $\mathcal{H}$ a tuple of indices (structure and facet). This allows points striking the same facet to be grouped into a point cluster. As shown in Figure~\ref{fig:point-cluster}, a \textbf{point cluster} with the index $c$ (visualized as borderless circles of the same color) comprises all points from a single scan that hit the same BIM facet. This creates an explicit link between BIM surface facets and point clusters. Given their similar planar nature, both the BIM facets and the point clusters are modeled as planes. Let $\{\pi_f\}$ be the set of analytically defined planes for the BIM facets, and $\{\pi_c\}$ be the set of planes for the point clusters. A plane $\pi_c$ for the point cluster  is parameterized by the point cluster's mean vector $\boldsymbol{\mu}_c$ (which defines a point on the plane) and the covariance matrix $\boldsymbol{\Sigma}_c$ (which defines the plane's normal).

Given $C$ point clusters and $F$ facets from a sequence of $K$ poses and $N$ BIM structures, the number of association pairs equals the number of point clusters {$C$.} The number of clusters associated with a specific facet $f$ is $C_f$, where  $C=\sum_{f=1}^{F}C_f$. These associations create two distinct constraints for trajectory optimization: (1) inter-scan consistency: all point clusters from different scans that are associated with the same facet must be geometrically consistent with each other; (2) BIM-scan consistency: each point cluster must align with its corresponding BIM surface facet.
Incorporating both constraints enhances the system's robustness. Relying solely on constraints from often-simplified BIM structures can yield incomplete constraints and suboptimal results. The combination of inter-scan and BIM-scan constraints mitigates this issue and corrects accumulated drift by using the BIM model as a global reference. To balance these constraints, the composite residual is decomposed as follows (for clarity, $\delta(k,n)$ and $\mathbb{I}_n$ are omitted):
\begin{equation}
        r(\mathcal{Z}, \mathcal{B}) = \sum_{f=1}^F \overbrace{r(\{\pi_c\}_{1:C_f})}^{(1)\,\mathcal{E}_\text{scan}:\,\text{Inter-Scan}} 
        + \sum_{f=1}^F \sum_{c=1}^{C_f} \overbrace{r(\pi_c, \pi_f).}^{(2)\,\mathcal{E}_\text{bim}:\,\text{BIM-Scan}} 
\end{equation}

This formulation separates the residual into two components: (1) $\mathcal{E}_\text{scan}$ that enforces inter-scan consistency by imposing a unified constraint on all point clusters $\{\pi_c\}_{1:C_f}$ associated with the same facet $f$. Because this is a single constraint involving multiple clusters, the residual cannot be separated per cluster. (2) $\mathcal{E}_\text{bim}$ that enforces BIM-scan consistency by evaluating the alignment between each point cluster $\pi_c$ and its corresponding BIM facet $\pi_f$. The following sections detail the residual formulations for these two constraints.

\subsubsection{Inter-scan consistency:} 
\label{sec:inter-scan-factor}
To formulate the constraints among the $C_f$ point clusters associated with the same facet $f$, we design a multi-way factor, with the residual representation as illustrated in Figure~\ref{fig:point-cluster}. Inspired by~\cite{Huang2021RAL}, we frame the bundle adjustment of point clusters as an eigenvalue minimization problem. For a point cluster $c$, the point distribution can be mostly captured by a mean vector $\boldsymbol{\mu}_c$ and a covariance matrix $\mathbf{\Sigma}_c$. The optimization objective $\mathcal{E}_\text{scan}$ is:
\begin{align}
        \mathcal{E}_\text{scan} = & \sum_{c=1}^{C_f} \lambda_{c}^1\big\|\boldsymbol{n}_g^\top (\mathbf{R}_k\mathbf{Q}_c \ \boldsymbol{e}_x)\big\|^2 + \sum_{c=1}^{C_f} \lambda_{c}^2\big\|\boldsymbol{n}_g^\top (\mathbf{R}_k\mathbf{Q}_c \ \boldsymbol{e}_y)\big\|^2  \notag \\
        + & \sum_{c=1}^{C_f} \|\boldsymbol{n}^{\top}_g(\mathbf{R}_{k}\boldsymbol{\mu}_c+\boldsymbol{t}_k -\boldsymbol{\mu}_g)\|^2,
    \label{eq:inter-scan}
\end{align}
where $\mathbf{R}_k$ is the rotation matrix at pose $x_k$. For each cluster $c$, the covariance matrix is decomposed via SVD as  $\mathbf{\Sigma}_{c}=\mathbf{Q}_c^{\top}\Lambda_c\mathbf{Q}_c$, where $\Lambda_c=\text{diag}(\lambda^1_{c}, \lambda^2_{c}, \lambda^3_{c})$,  $\boldsymbol{e}_x=[1,0,0]^{\top}$, and $\boldsymbol{e}_y=[0,1,0]^{\top}$. The terms $\boldsymbol{\mu}_g$ and $\boldsymbol{n}_g$ are derived from the aggregated distribution of all $C_f$ point clusters. The global mean $\boldsymbol{\mu}_g$ is the weighted sum of the individual cluster means  $\boldsymbol{\mu}_c$. The global normal vector $\boldsymbol{n}_g$ comes from the SVD of the aggregated covariance matrix $\mathbf{\Sigma}_g$ that is calculated as:
$\mathbf{\Sigma}_g = \sum_{c=1}^{C_f}(\mathbf{\Sigma}_c+\mathbf{\Sigma}_{\boldsymbol{\mu}_c})$, where $\quad \boldsymbol{\Sigma}_{\boldsymbol{\mu}_c}= (\boldsymbol{\mu}_c-\boldsymbol{\mu}_g)(\boldsymbol{\mu}_c-\boldsymbol{\mu}_g)^{\top}$. 

Eq.~\ref{eq:inter-scan} is designed to reformulate inter-scan consistency problem into the {thickness} of the aggregated point distribution along $\boldsymbol{n}_g$. This objective is achieved by iteratively evaluate the similarity between eigenvector related to $\lambda_c^3$ of point clusters and $\boldsymbol{n}_g$ and update until they are almost the same. The optimization proceeds as two-fold iteratively: (1) hold $\boldsymbol{\mu}_g$ and  $\boldsymbol{n}_g$ fixed to update all cluster parameters $\boldsymbol{\mu}_c$ and  $\mathbf{R}_k$ and (2) use the updated cluster parameters to recompute $\boldsymbol{\mu}_g$ and  $\boldsymbol{n}_g$. This process repeats until convergence, tightly coupling the point clusters from multiple scans through their associated BIM structures.

\subsubsection{BIM-scan consistency}
\label{sec:bim-scan-factor}
This factor enforces consistency between associated BIM structures and point clusters. Using the synthetic points sampled from BIM surfaces via multi-hit ray casting, the BIM-scan residual is defined by the point-to-surface distance between real-world scan points and the local BIM surfaces. This is illustrated on the right side of Figure~\ref{fig:multi-hit}. The residual is formulated as:
\begin{equation}
    \mathcal{E}_\text{bim} =\sum_{i=1}\|{\boldsymbol{n}^{\prime}_{i}}^{\top}(\mathbf{R}_k\boldsymbol{p}_{i} + \boldsymbol{t}_k - \boldsymbol{p}^{\prime}_{i})\|^2 \ .
    \label{eq:bim-obs}
\end{equation}
Here, each point index by $i$ in a point cluster is associated with a BIM facet. The variables  $\boldsymbol{p}^{\prime}_{i}$ and $\boldsymbol{n}^{\prime}_{i}$ are the associated synthetic hit point and its normal vector, respectively. As shown in Figure~\ref{fig:point-cluster}, the distance for a real-world point $\boldsymbol{p}_i$ is calculated relative to the plane defined by the synthetic point $\boldsymbol{p}^{\prime}_{i}$ and its normal $\boldsymbol{n}^{\prime}_{i}$.

\subsubsection{Factor graph representation}
\label{sec:factor-graph}
The previously defined constraints and their residual formulations deeply integrate the BIM model into pose estimation. The resulting factor graph for trajectory optimization is illustrated in Figure~\ref{fig:factor-graph}. It incorporates the following factors: (1) odometry factor that provides motion priors from the front-end module; (2) inter-scan consistency factor (Section~\ref{sec:inter-scan-factor}) that unifies point clusters from sequential scans into a single optimization target, represented by the dark gray region; (3) BIM-scan consistency factor (Section~\ref{sec:bim-scan-factor}) that aligns scans with the BIM model. It first transform and accumulate scans into a temporary sub-map and then uses multi-hit ray casting to generate associations between this sub-map and the BIM structures, as shown in the light gray region. Combined with a prior on the initial state, these factors optimize the trajectory and align it with the BIM coordinate system.

\begin{figure}[t]
    \centering
    \includegraphics[width=1.0\linewidth]{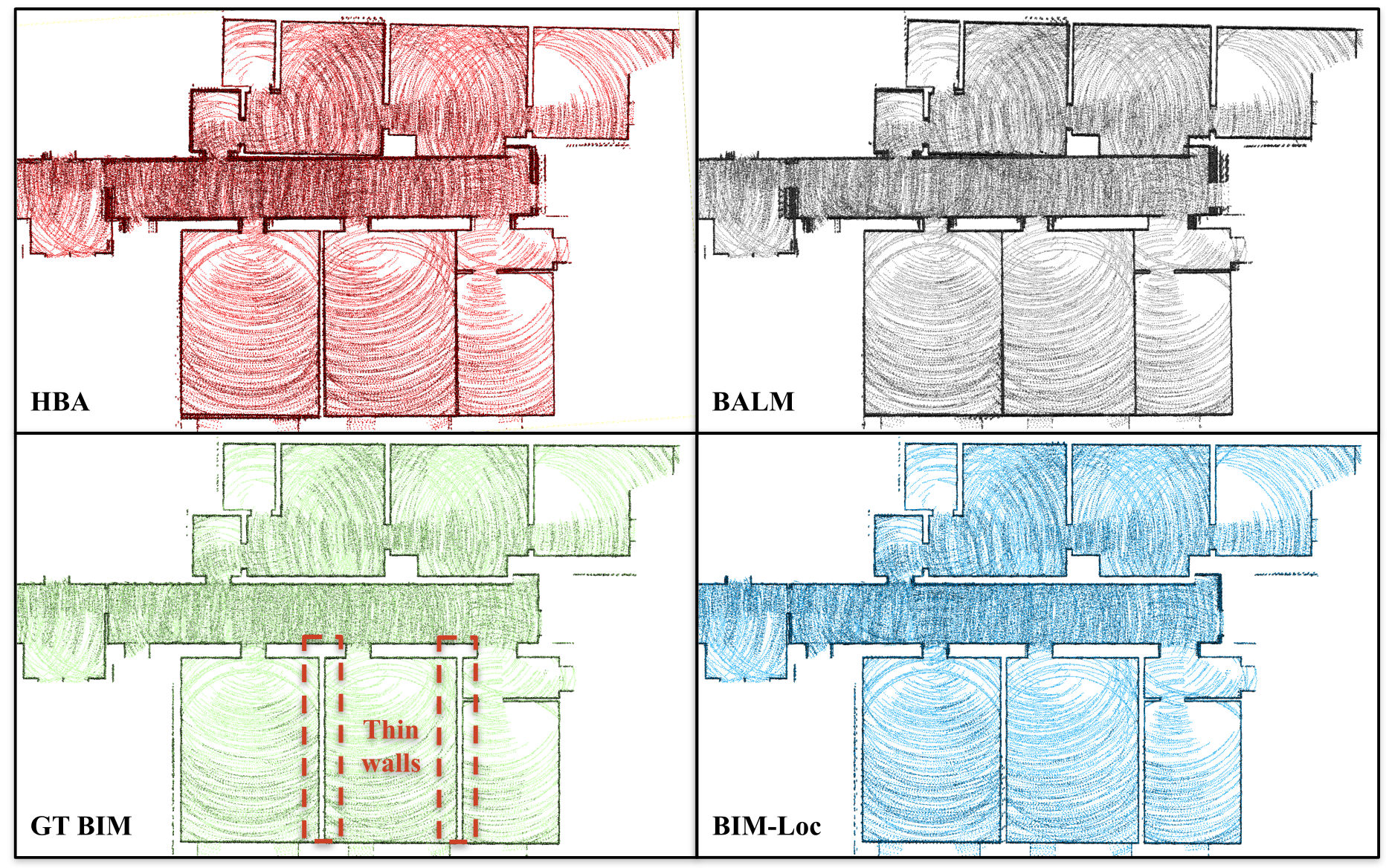}
    \caption{
Challenge of voxel-based data association with thin wall structures. Walls measuring {7.5} to {10}{cm} in thickness are comparable to typical voxel resolutions used in large-scale buildings, leading to ambiguous data associations.
    }
    \label{fig:thin-wall}
\end{figure} 
Furthermore, integrating BIM models through these constraints provides richer geometric knowledge of the environment. As shown in Figure~\ref{fig:thin-wall}, indoor scenes often contain thin walls (bottom left). {Because our method uses the explicit boundaries defined in the BIM, the boundaries of these thin walls can be clearly recovered by accumulated point clouds.} In contrast, methods like HBA~\citep{Liu2023RAL} and BALM2~\citep{Liu2023TRO} fail to distinctly separate the two sides of a thin wall.

\begin{figure*}[t]
    \centering
    \includegraphics[width=1.0\linewidth]{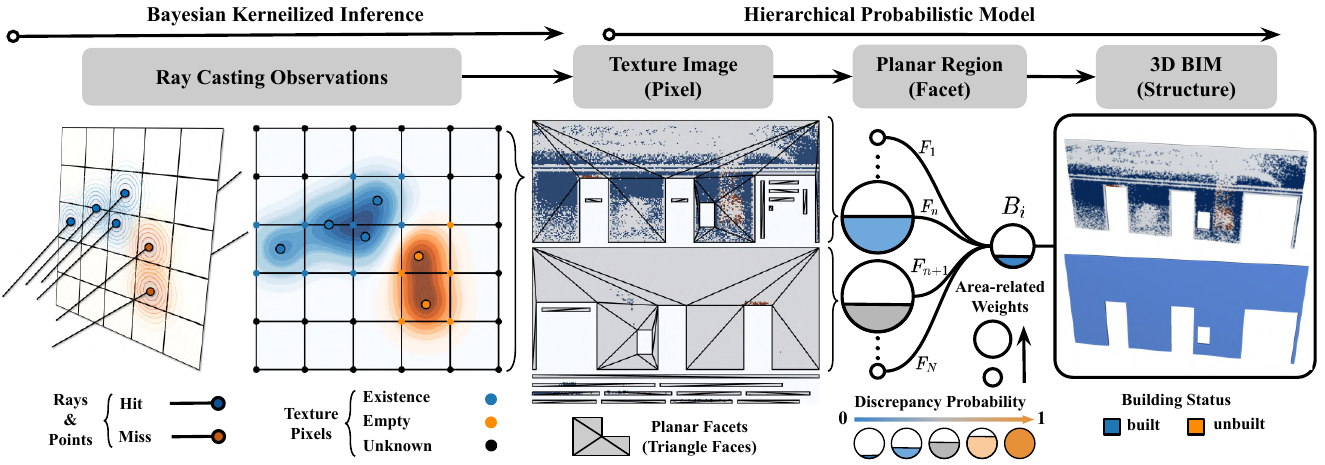}
    \caption{
    Hierarchical Bayesian model for online discrepancy detection. The model infers structural discrepancies from sparse ray-casting observations using a multi-stage Bayesian inference pipeline. Sparse observations are first processed through kernelized Bayesian inference to predict pixel status in texture images, then propagated through pixel-facet-structure hierarchies to estimate BIM structure discrepancy indicators.
    }
    \label{fig:hierarchical-model}
\end{figure*}

\subsection{Discrepancy Detection}
\label{sec:discrepancy-identification} 
This section addresses the subproblem of detecting discrepancies in BIM structures. As shown in Eq.~\ref{eq:problem-costs}, the status of each structure is represented by a discrepancy indicator, denoted here as $\mathbb{I}_\text{S}$ (the structure index $n$ is omitted for clarity and we use subscript $S$ to emphasize its affiliation within structure-level representation). This structure-level discrepancy indicator is a discrete random variable following a Bernoulli distribution: $\mathbb{I}_\text{S}\sim \text{Bern}(p;\theta_\text{S})={\theta_\text{S}}^p(1-\theta_\text{S})^{1-p}$, where $p \in \{0, 1\}$. The hyper-parameter $\theta_\text{S}$ is the probability that a discrepancy exists ($\mathbb{I}_\text{S}=1$) when comparing the BIM structure against actual sensor observations, e.g., $\mathbb{I}_\text{S}=1$ when the discrepancy exists.

The goal of discrepancy detection is to estimate the posterior distribution $p(\theta_\text{S} \mid \mathcal{Z})$. However, two challenges arise: sparse point cloud observations cannot be directly used for structure-level identification, and clear comparison criteria are lacking. To address these issues, we propose a hierarchical continuous mapping method in the surface texture spaces of BIM structures, as illustrated in Figure~\ref{fig:hierarchical-model}. This strategy converts sparse ray-casting observations into real-time estimates of the discrepancy status $\mathbb{I}_\text{S}$.

\subsubsection{Hierarchical Discrepancy Modeling}
\label{sec:hierarchical-model}
To bridge this representation gap, we design a probabilistic model based on the hierarchical architecture of BIM models: structure, facet, texture pixel, and {observations $\mathcal{Z}$}. The prediction objective is decomposed as follows:
\begin{equation}
    p(\theta_\text{S}\mid\mathcal{F}, \mathcal{I}, \mathcal{Z}) \propto p(\theta_\text{S}\mid\mathcal{F})\cdot p(\mathcal{F}\mid\mathcal{I}) \cdot p(\mathcal{I}\mid\mathcal{Z})\ , 
    \label{eq:hierarchical-model}
\end{equation}
where $\mathcal{F}\coloneq\{\mathbb{I}_\text{F}\}$ and {$\mathcal{I}\coloneq\{\mathbb{I}_\text{I}\}$} are two sets of auxiliary discrete random variables. $\mathbb{I}_\text{F}$ follows a Bernoulli distribution, $\mathbb{I}_\text{F}\sim \text{Bern}(p;\theta_\text{F})$, with hyperparameter $\theta_F$ representing the probability that a BIM surface facet exhibits a discrepancy ($\mathbb{I}_\text{F}=1$). {$\mathbb{I}_\text{I}$ is pixel-wise discrepancy indicator, which will be defined in Equation~\ref{eq:pixel-indicator}.} Equation~\ref{eq:hierarchical-model} forms a three-layer hierarchical model with a chain structure from observations, texture pixels, facets to building structures: $\mathcal{Z}\rightarrow\{\mathbb{I}_\text{I}\}\rightarrow\{\mathbb{I}_\text{F}\}\rightarrow\mathbb{I}_\text{S}$. Two relationships are key in this chain: (1) the structure-facet relationship $\{\mathbb{I}_\text{F}\}\rightarrow\mathbb{I}_\text{S}$ and (2) the facet-pixel relationship $\{\mathbb{I}_\text{I}\}\rightarrow\mathbb{I}_\text{F}$. These relationships are defined as follows:

\textbf{Structure-facet relationship} between the structure indicator $\mathbb{I}_\text{S}$ and its facet indicators $\mathbb{I}_F$ defined via the hyperparameter $\theta_S$:
\begin{equation}
    \mathbb{I}_\text{S}\sim Bern(\theta_\text{S}), \, \theta_\text{S}=\bigcup_\mathcal{F}(w_\text{F}\cdot\mathbb{I}_\text{F}) \ ,
    \label{eq:structure-facet-relation}
\end{equation}
where $\theta_\text{S}$ is the weighted union of the facet discrepancy indicators in $\mathcal{F}$. This formulation reflects the composition of a complete building structure from its facets. To prioritize larger facets, we define an area-based weight $w_\text{F} = {A_F}^2 / \sum{A_\text{F}}^2$, where $A_F$ is the area of a facet. This ``larger-area, higher-weight'' strategy focuses the model's updates on more significant surfaces.

\textbf{Facet-pixel relationship} leverages the surface texture projection from Section~\ref{sec:multi-hit}, where BIM facets are projected onto a texture image with pre-computed barycentric coordinates. The connection is established through a model averaging of pixels:
\begin{align}
    \mathbb{I}_\text{F}\sim Bern(\theta_\text{F}), \, \theta_\text{F}=\frac{1}{N_\text{pixel}}\sum\mathbb{I}_\text{I} \ , 
    \label{eq:facet-pixel-relation}
\end{align}
where $N_\text{pixel}$ is the number of pixels in the facet and $\mathbb{I}_\text{I}$ is the pixel-wise discrepancy indicator. With these definitions, the problem of detecting discrepancies in a BIM structure is equivalent to identifying the status of its corresponding texture pixels, which form the supporting set for each planar facet. The next section defines the pixel discrepancy indicator and its connection to point cloud observations.

\subsubsection{Bayesian Kernelized Inference}
\label{sec:kernelized-inference}
Building on the hierarchical probabilistic model, this section describes a continuous texture mapping strategy using sparse point cloud observations to establish the relationship  $\mathcal{Z}\rightarrow\mathbb{I}_{\text{I}}$  between points and texture pixels. This mapping must fulfill two requirements: (1) model the pixel discrepancy indicator $\mathbb{I}_\text{I}$ to determine discrepancies in each pixel's local surface neighborhood; and (2) support incremental updates for online sequential measurements.

As noted in Section~\ref{sec:multi-hit}, laser rays are organized into hit ($\mathcal{H}$) and miss ($\mathcal{M}$) sets, with point positions defined by UV coordinates. However, unlike the uniform grid of a texture image, ray intersections occur at arbitrary coordinates and provide only sparse coverage of facet surfaces. To address this, we apply Bayesian kernelized inference, inspired by methods in occupancy voxel mapping~\citep{Doherty2019TRO} and terrain traversability mapping~\citep{Shan2018PMLR}. Specifically, we define a series of 2D continuous spaces covering the planar facets of each BIM structure; these spaces are proper subsets of the texture image space. Each pixel within a facet serves as an anchor, representing the local status of its neighborhood through the pixel discrepancy indicator:
\begin{equation}
    \mathbb{I}_\text{I} \sim Beta(\alpha_\text{I},\beta_\text{I}) \ , 
    \label{eq:pixel-indicator}
\end{equation}
where a Beta distribution models the accumulated status, parameterized by hyper-parameters  $\alpha_\text{I}$ and  $\beta_\text{I}$. For each ray intersection with status  $\sigma$ (0 for miss, 1 for hit) and barycentric coordinates $(u, v)$, a texture-wrapping lookup table identifies the neighborhood pixels (Section~\ref{sec:multi-hit}). Given sequential observations, the hyper-parameters of affected anchor pixels are updated incrementally:
\begin{equation}
    \left\{
        \begin{aligned}
            \alpha_\text{I} &\leftarrow \alpha_\text{I} + \sum_{x^{\prime}\in\mathcal{N}(x)} k(x, x^{\prime})\cdot(1-\sigma) \ ,  \\
            \beta_\text{I} &\leftarrow \beta_\text{I} + \sum_{x^\prime\in\mathcal{N}(x)} k(x, x^\prime)\cdot\sigma \ , 
        \end{aligned}
    \right.
    \label{eq:pixel-update}
\end{equation}
where each anchor pixel $x$ is updated using all neighborhood measurements $x^\prime \in \mathcal{N}(x)$. The kernel function $k(\cdot, \cdot)$ weights the influence of each measurement. We directly adopt the sparse kernel from~\cite{Doherty2019TRO} to incorporate measurements from the spatial neighborhood with three reasons:
{\emph{1. Computational tractability}: its compact support restricts computation to local neighborhoods, avoiding the intractable cost of dense RBF kernels over all historical observations. \emph{2. Physical locality}: the hard cutoff ensures query points are evaluated only against locally adjacent surfaces, preventing false correspondences and physically implausible inference. 3. \emph{Online incremental update}: only nearby pixels contribute and new measurements are incorporated through exact recursive Bayesian updates to local $(\alpha_{\text{I}}, \beta_{\text{I}})$ without revisiting the full history.}

\subsubsection{Discrepancy Status Propagation}
\label{sec:map-inference}
The hierarchical model and kernelized inference establish a complete chain for discrepancy propagation: $\mathcal{Z}\rightarrow\{\mathbb{I}_\text{I}\}\rightarrow\{\mathbb{I}_\text{F}\}\rightarrow\mathbb{I}_\text{S}$. At level of observations and texture images, laser rays update anchor pixels via Eq.~\ref{eq:pixel-update}. We characterize each pixel's state using the mean and variance of its discrepancy indicator $\mathbb{I}_\text{I}$. Given its Beta distribution, these statistics are:
\begin{equation}
    \left\{
    \begin{aligned}
        \mathbb{E}[\mathbb{I}_\text{I}] &= \frac{\alpha_\text{I}}{\alpha_\text{I} +\beta_\text{I}} \ , \\
        \mathbf{Var}[\mathbb{I}_\text{I}] &= \frac{\alpha_\text{I}\beta_\text{I}}{(\alpha_\text{I}+\beta_\text{I})^2(\alpha_\text{I} +\beta_\text{I}+1)} \ . 
    \end{aligned}
    \right.
    \label{eq:pixel-inference}
\end{equation}

These pixel-level estimates are then propagated upward. For the facet-level indicator $\mathbb{I}_\text{F}$, which also follows a Bernoulli distribution, the estimates are derived by model averaging over the facet's pixels:
\begin{equation}
    \left\{
    \begin{aligned}
        \mathbb{E}[\mathbb{I}_\text{F}] &= \frac{1}{N_\text{pixel}}\sum\mathbb{E}[\mathbb{I}_\text{I}] \ , \\
        \mathbf{Var}[\mathbb{I}_\text{F}] &= \mathbb{E}[\mathbb{I}_\text{F}] -\frac{1}{{N_\text{pixel}}^2}\sum\mathbf{Var}[\mathbb{I}_\text{I}]- (\mathbb{E}[\mathbb{I}_\text{F}])^2. \\
    \end{aligned}
    \right.
    \label{eq:facet-inference}
\end{equation}

Finally, the structure-level estimates are computed from the facet-level indicators using the weighted union from Eq.~\ref{eq:structure-facet-relation}:
\begin{equation}
    \left\{
    \begin{aligned}
        \mathbb{E}[\mathbb{I}_\text{S}] &= \sum{w}_\text{F}\cdot\mathbb{E}[\mathbb{I}_\text{F}] \ , \\ 
        \mathbf{Var}[\mathbb{I}_\text{S}] &= \mathbb{E}[\mathbb{I}_\text{S}] - \sum{w}_\text{F}^2\cdot\mathbf{Var}[\mathbb{I}_\text{F}] - (\mathbb{E}[\mathbb{I}_\text{S}])^2 \ . \\
    \end{aligned}
    \right.
    \label{eq:structure-inference}
\end{equation}

\begin{figure}[t]
    \centering
    \includegraphics[width=1.0\linewidth]{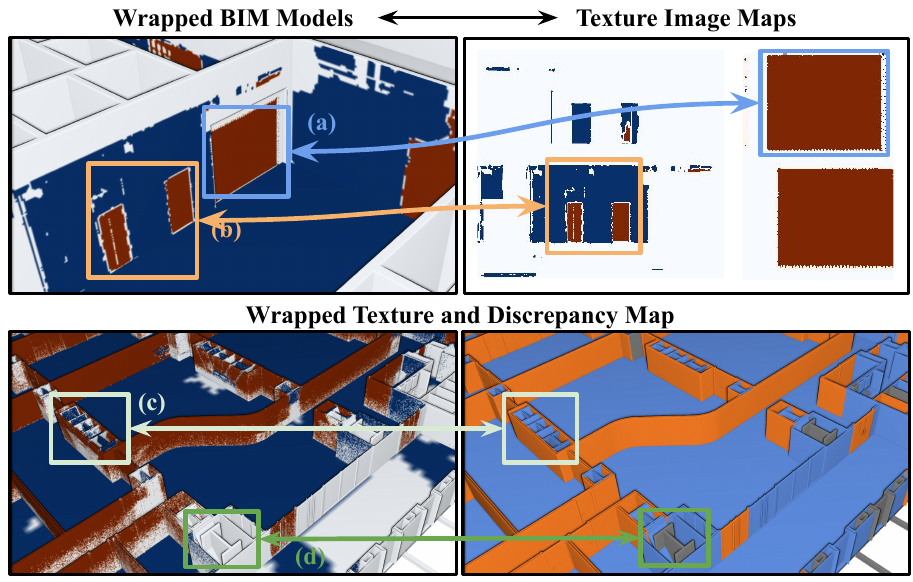}
    \caption{
    Texture mapping visualization for discrepancy detection. Cases (a) and (b) show wrapped BIM models aligned with texture maps, highlighting discrepancies from elevator and roller doors. Cases (c) and (d) show H-shape beam detection behind unfinished and finished walls. Color coding indicates alignment status: \textcolor{blue}{blue} for consistent, \textcolor{orange}{orange} for discrepant, and \textcolor{gray}{gray} for unknown regions.
    }
    \label{fig:discrepancy-visualization}
\end{figure}

The definitions in Eq.~\ref{eq:facet-inference} and Eq.~\ref{eq:structure-inference} enable the propagation of discrepancy estimates from point cloud observations across all levels, from individual pixels to entire BIM structures. The final discrepancy indicator $\hat{\mathbb{I}}$ for a facet or structure is determined by:
\begin{equation}
    \hat{\mathbb{I}} \coloneqq
    \left\{
    \begin{aligned}
        &1 \ , \text{if} \,\,\mathbf{E}[\mathbb{I}] \geq \gamma^{+}_\mu \ , \mathbf{Var}[\mathbb{I}]\leq \gamma_{\sigma} \, \text{(consistent)} \\
        &0 \ , \text{if} \,\,\mathbf{E}[\mathbb{I}] \leq \gamma^{-}_\mu \ , \mathbf{Var}[\mathbb{I}]\leq \gamma_{\sigma} \, \text{(discrepant)} \\ 
        &0 \ , \text{otherwise} \, \text{(unknown)} \\
    \end{aligned}
    \right. \ , 
    \label{eq:indicator}
\end{equation}
where the hyper-parameters $\gamma^{+}_{\mu}$, $\gamma^{-}_{\mu}$, and $\gamma_{\sigma}$ control the sensitivity of the detection.

For a BIM structure, $\hat{\mathbb{I}}_{S}=1$ indicates it exists in the actual scene. When $\hat{\mathbb{I}}_{S}=0$, two scenarios are possible: (1) If the variance $\mathbf{Var}(\mathbb{I}_S)$ is below $\gamma_\sigma$, the structure is discrepant (i.e., it differs from the BIM model). (2) Otherwise, it is classified as unknown, indicating insufficient data. 

The effectiveness of this module is shown in Figure~\ref{fig:discrepancy-visualization}. For instance, in case (c), the module correctly distinguishes completed H-shape beams from nearby walls, while In case (d), a square beam is identified as completed, while the H-beam inside it is labeled ``unknown'' due to lack of observations.

Each indicator in the hierarchy serves a distinct purpose: $\mathbb{I}_\text{I}$ manages fine-grained, pixel-level discrepancy checking, revealing surface details;
$\mathbb{I}_\text{F}$ facilitates data association for trajectory optimization (Section~\ref{sec:bim-factor}), and $\mathbb{I}_\text{S}$ determines the existence of a BIM structure for high-level discrepancy reporting.

\section{Experiments}
\label{sec:experiments}
Our BIM-Loc method is evaluated across three distinct environmental settings: a BIM-robot simulation platform, the open-source SLABIM dataset~\citep{Huang2025ICRA}, and a dataset collected from a real-world construction site~\citep{Zhang2024AIC}. Each setting serves as a separate benchmark, with comparisons made against several state-of-the-art approaches. All experiments were executed on a Mini-PC integrated with the sensor suite, utilizing an Intel\textregistered{} Core\texttrademark{} i9-12900 CPU for computation. Notably, all evaluated algorithms rely solely on CPU-based processing, with no NVIDIA GPU acceleration required. As depicted in Figure~\ref{fig:overall-framework}, the inputs to BIM-Loc consist of sensor data sequences (including scans and IMU measurements) and BIM models. For other algorithms requiring point-based prior maps, these were uniformly sampled from the surfaces of the same as-designed BIM models used by our method.

\begin{figure*}[t]
    \centering
    \includegraphics[width=1.0\linewidth]{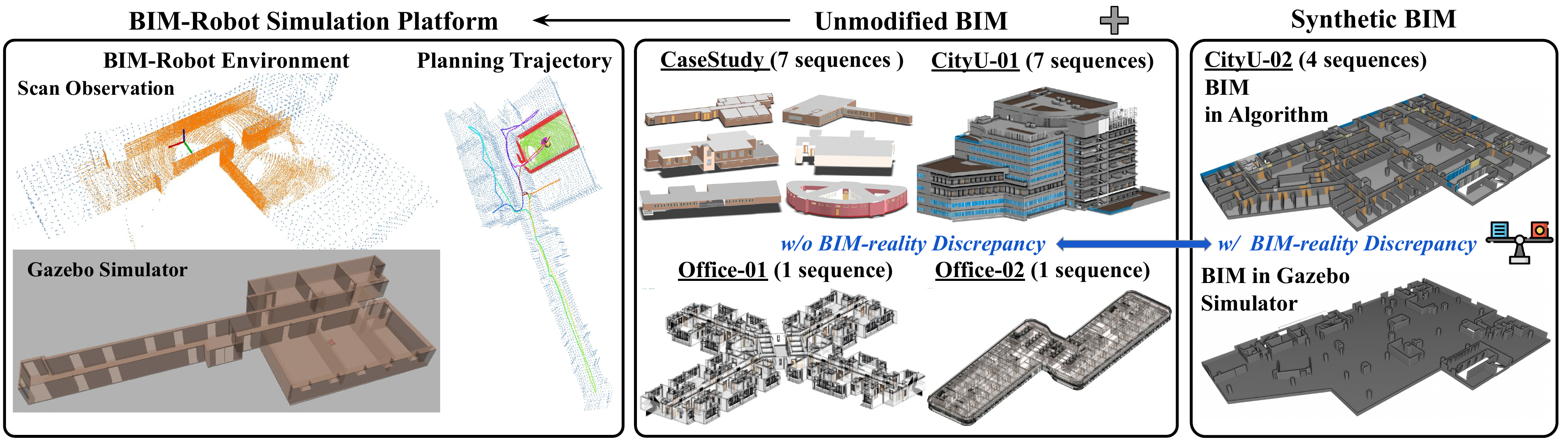}
    \caption{
    Simulation benchmark structure, comprising a Gazebo simulator with BIM models and an exploration algorithm. Five BIM model groups generate test data: four unmodified models (CaseStudy, CityU-01, Office-01, Office-02) evaluate localization {without BIM-reality discrepancies}, while a synthetic model (CityU-02) focuses explicitly on such discrepancies.
    }
    \label{fig:sim-dataset}
\end{figure*}

To comprehensively assess BIM-Loc's performance, five evaluation metrics are employed. For trajectory accuracy, we use Absolute Trajectory Error (ATE), calculating the Root Mean Square Error (RMSE) for both translation and rotation components:
\begin{equation}
        \text{ATE} =\left\{
        \begin{aligned}
            & \|\boldsymbol{t} - \boldsymbol{t}_{\text{gt}}\| \ , \quad\text{\textit{translation error}} \\
            & \|\mathbf{R}^{-1}_{\text{gt}}\mathbf{R} - \mathbf{I}_{3\times 3}\|_\text{F} \ , \quad\text{\textit{rotation error}} \\ 
        \end{aligned}
        \right. \ . \\
    \label{eq:ate}
\end{equation}

Additionally, Mean Map Entropy (MME)~\citep{Kornilova2021ECMR, Hu2025RAL} is used to evaluate the local spatial consistency of the aggregated point cloud map $\mathcal{P}$ and relative pose errors. MME computes the average smallest eigenvalues of the local covariance matrix of each point $\boldsymbol{p}_i$ in the aggregated map, requiring only the point cloud data itself:
\begin{equation}
    \text{MME} = -\frac{1}{|\mathcal{P}|}\sum_{i=1}^{|\mathcal{P}|}\lambda_{\min}(\boldsymbol{p}_i) \ .
    \label{eq:mme}
\end{equation}

To account for variations in MME values across different BIM models, we normalize the results to get the MME ratio for fair comparison, which is computed as $\text{MME}_{\text{agg}}/\text{MME}_{\text{bim}}\in [0, 1]$, which represents the ratio between the MME of the aggregated map and that of the sampled BIM surface map.

Beyond trajectory errors, we evaluate BIM-observation consistency using another two metrics. The scan-to-BIM distance metric assesses per-scan consistency by computing distances between points and BIM surfaces. To reduce noise impact, only distances below a predefined threshold (0.2~m in all experiments) are considered for RMSE computation. For global consistency evaluation, we employ the Wasserstein Distance (WD) metric proposed by~\cite{Hu2025RAL}. The ground truth map and estimated aggregated map are first voxelized. The voxel-wise WD error is computed according to the distribution of the inside points:
\begin{equation}
    \text{WD} = \sqrt{ 
    \begin{aligned}
        & \|\boldsymbol{\mu}_\text{v}^{\text{gt}} - \boldsymbol{\mu}_\text{v}^{\text{est}}\|^2 + \\ 
        &\text{Tr}\Big(\mathbf{\Sigma}_\text{v}^\text{gt}+\mathbf{\Sigma}_\text{v}^{\text{est}}- 2\sqrt{(\mathbf{\Sigma}_\text{v}^\text{est})^{\frac{1}{2}}\mathbf{\Sigma}_\text{v}^\text{gt}(\mathbf{\Sigma}_{\text{v}}^\text{est})^{\frac{1}{2}}}\Big)\\
    \end{aligned}
    } \ ,
\end{equation}
where $\boldsymbol{\mu}_\text{v}$ and $\mathbf{\Sigma}_\text{v}$ are mean vector and covariance matrix of points belonging to the same 3D voxel $\text{v}$ in both ground truth map ($\text{gt}$) and estimated global map ($\text{est}$), respectively. $\text{Tr}(\cdot)$ denotes the trace operation. The global consistency is evaluated by the distributions of WD errors. For discrepancy detection performance, we report precision, recall, and F1-score metrics. Additionally, we visualize the discrepancy detection results to qualitatively validate the performance of BIM-Loc.

{The BIM models used in the work are all with LOD~300 (i.e., Levels of Details) and come from the International Foundation Classes (IFC), which is a popular in AEC fields for efficient building data sharing. To interpret the BIM model into the objects of building structures in robot application and simulation platforms, a series of tool chain is designed focusing on two functionalities: 1. transfer the raw IFC files into meshes of building elements, and 2. generate URDF files for usage in simulation platforms. In BIM/IFC files, the building structures are organized. Please refer to Appendix~\ref{sec:bim-preprocess} for more information.}

\subsection{Simulation Platform}
\label{sec:sim-platform}
 The initial evaluation targets indoor environments that are free of temporary objects not captured in the BIM models. To facilitate this, we developed a BIM-robot simulation platform built upon an autonomous exploration framework~\citep{Cao2022ICRA}. As illustrated in Figure~\ref{fig:sim-dataset}, BIM models are converted into \textit{.stl} format and imported into the Gazebo simulator~\citep{Aguero2015Gazebo}. A mobile robot equipped with a Velodyne VLP-16 LiDAR sensor navigates through the virtual building along trajectories generated by an exploration algorithm~\citep{Cao2021RSS}, producing synthetic LiDAR scans. The resulting synthetic data sequences, along with the BIM models, serve as inputs for all evaluated algorithms.

\begin{table*}[t]
    \caption{
    Simulation benchmark results for ATE and scan-to-BIM distances. Column subscripts: $\boldsymbol{t}$ = ATE translation [m], $\boldsymbol{\theta}$ = ATE rotation [deg], $d$ = scan-to-BIM distance [m]. Lower values indicate better performance. Optimal results are in bold and next-best are underlined.
    }
    \label{tab:ape-rmse-results}
    \centering
    \scriptsize
    \renewcommand{\arraystretch}{1.3}
    \setlength{\tabcolsep}{2.5pt}
    \begin{tabular*}{\textwidth}{@{\extracolsep\fill} lccc|ccc|ccc|ccc|ccc|ccc@{\extracolsep\fill}}
        \toprule[1.2pt]
        \multicolumn{1}{c}{\multirow{2}{*}{\footnotesize\textbf{Sequence}}} &
        \multicolumn{3}{c}{\footnotesize\textbf{BIM-Loc}} &
        \multicolumn{3}{c}{\footnotesize\textbf{DLO}} &
        \multicolumn{3}{c}{\footnotesize\textbf{HBA}} &
        \multicolumn{3}{c}{\footnotesize\textbf{BALM2}} &
        \multicolumn{3}{c}{\footnotesize\textbf{HDL}} &
        \multicolumn{3}{c}{\footnotesize\textbf{SC-PGO}} \\
        & $\boldsymbol{t}$ & $\boldsymbol{\theta}$ & $d$ & $\boldsymbol{t}$ & $\boldsymbol{\theta}$ & $d$ & $\boldsymbol{t}$ & $\boldsymbol{\theta}$ & $d$ & $\boldsymbol{t}$ & $\boldsymbol{\theta}$ & $d$ & $\boldsymbol{t}$ & $\boldsymbol{\theta}$ & $d$ & $\boldsymbol{t}$ & $\boldsymbol{\theta}$ & $d$ \\
        \midrule
        CaseStudy-01   & $\textbf{0.011}$ & $\textbf{0.147}$ & $\textbf{0.009}$ & $0.174$ & $\underline{0.847}$ & $0.158$ & $0.181$ & $0.766$ & $0.163$ & $0.237$ & $1.595$ & $0.222$ & $\underline{0.157}$ & $2.469$ & $\underline{0.142}$ & $0.203$ & $1.916$ & $0.189$ \\
        CaseStudy-02-1 & $\textbf{0.029}$ & $\textbf{0.719}$ & $\textbf{0.015}$ & $0.188$ & $1.098$ & $0.168$ & $\underline{0.173}$ & $\underline{0.900}$ & $\underline{0.155}$ & $0.193$ & $1.210$ & $0.160$ & $0.231$ & $7.253$ & $0.241$ & $0.193$ & $1.836$ & $0.182$ \\
        CaseStudy-02-2 & $\textbf{0.034}$ & $\textbf{0.187}$ & $\textbf{0.011}$ & $0.260$ & $1.326$ & $0.219$ & $0.240$ & $1.228$ & $0.205$ & $0.209$ & $\underline{0.915}$ & $\underline{0.166}$ & $0.208$ & $3.654$ & $0.199$ & $\underline{0.183}$ & $2.130$ & $0.183$ \\
        CaseStudy-03   & $\textbf{0.009}$ & $\textbf{0.080}$ & $\textbf{0.008}$ & $0.056$ & $0.257$ & $0.053$ & $\underline{0.039}$ & $\underline{0.223}$ & $\underline{0.037}$ & $0.052$ & $0.504$ & $0.051$ & $0.055$ & $1.962$ & $0.108$ & $0.111$ & $1.302$ & $0.123$ \\
        CaseStudy-04   & $\textbf{0.007}$ & $\textbf{0.093}$ & $\textbf{0.008}$ & $0.023$ & $0.143$ & $0.022$ & $\underline{0.016}$ & $\underline{0.123}$ & $\underline{0.016}$ & $0.031$ & $0.193$ & $0.030$ & $1.180$ & $1.676$ & $0.862$ & $0.145$ & $0.682$ & $0.116$ \\
        CaseStudy-05   & $\textbf{0.008}$ & $\textbf{0.124}$ & $\textbf{0.008}$ & $\textbf{0.008}$ & $0.179$ & $\underline{0.009}$ & $0.010$ & $\underline{0.175}$ & $0.011$ & $0.017$ & $0.864$ & $0.027$ & $0.301$ & $17.13$ & $0.381$ & $0.287$ & $1.969$ & $0.285$ \\
        CaseStudy-06   & $\textbf{0.043}$ & $\textbf{0.323}$ & $\textbf{0.023}$ & $\underline{0.065}$ & $\underline{0.921}$ & $\underline{0.057}$ & $0.087$ & $1.409$ & $0.075$ & $0.074$ & $1.046$ & $0.068$ & $0.194$ & $3.436$ & $0.232$ & $0.161$ & $2.058$ & $0.171$ \\
        CityU-01-F06   & $\textbf{0.052}$ & $\textbf{0.930}$ & $\textbf{0.014}$ & $0.354$ & $3.762$ & $0.314$ & $0.334$ & $3.558$ & $0.292$ & $1.161$ & $6.357$ & $0.363$ & $\underline{0.241}$ & $\underline{2.118}$ & $\underline{0.175}$ & $1.821$ & $4.530$ & $1.740$ \\
        CityU-01-F07   & $\textbf{0.008}$ & $\textbf{0.151}$ & $\textbf{0.006}$ & $0.471$ & $\underline{1.566}$ & $0.394$ & $\underline{0.448}$ & $3.084$ & $\underline{0.389}$ & $0.486$ & $1.723$ & $0.411$ & $0.574$ & $4.066$ & $0.508$ & $0.726$ & $2.079$ & $0.700$ \\
        CityU-01-F08   & $\textbf{0.203}$ & $\textbf{0.622}$ & $\textbf{0.153}$ & $\underline{1.320}$ & $\underline{1.701}$ & $\underline{0.932}$ & $1.340$ & $2.296$ & $0.948$ & $1.329$ & $2.249$ & $0.952$ & $8.689$ & $72.29$ & $6.967$ & $1.333$ & $3.380$ & $1.059$ \\
        CityU-01-F09   & $\textbf{0.012}$ & $\textbf{0.281}$ & $\textbf{0.007}$ & $0.443$ & $1.628$ & $0.403$ & $\underline{0.413}$ & $1.945$ & $0.372$ & $0.597$ & $2.366$ & $0.466$ & $\underline{0.185}$ & $2.192$ & $\underline{0.145}$ & $1.314$ & $2.385$ & $1.204$ \\
        CityU-01-F10   & $\textbf{0.014}$ & $\textbf{0.133}$ & $\textbf{0.007}$ & $0.386$ & $1.554$ & $0.327$ & $0.291$ & $\underline{1.133}$ & $0.253$ & $0.396$ & $1.742$ & $0.340$ & $\underline{0.103}$ & $2.014$ & $\underline{0.127}$ & $0.497$ & $3.182$ & $0.498$ \\
        CityU-01-F11   & $\textbf{0.013}$ & $\textbf{0.233}$ & $\textbf{0.007}$ & $0.356$ & $\underline{1.222}$ & $0.331$ & $0.292$ & $1.232$ & $0.271$ & $0.373$ & $1.431$ & $0.336$ & $\underline{0.111}$ & $2.000$ & $\underline{0.119}$ & $0.666$ & $3.266$ & $0.616$ \\
        CityU-01-F12   & $\textbf{0.021}$ & $\textbf{0.188}$ & $\textbf{0.008}$ & $\underline{0.129}$ & $\underline{0.975}$ & $\underline{0.128}$ & $0.132$ & $1.072$ & $0.131$ & $0.522$ & $1.140$ & $0.200$ & $0.403$ & $3.256$ & $0.272$ & $2.669$ & $10.71$ & $1.652$ \\
        Office-01      & $\textbf{0.018}$ & $\textbf{0.245}$ & $\textbf{0.009}$ & $0.265$ & $\underline{1.418}$ & $0.237$ & $0.231$ & $1.615$ & $0.207$ & $0.292$ & $1.558$ & $0.269$ & $\underline{0.155}$ & $3.823$ & $\underline{0.167}$ & $0.552$ & $3.640$ & $0.535$ \\
        Office-02      & $\textbf{0.037}$ & $\textbf{0.115}$ & $\textbf{0.007}$ & $0.383$ & $1.072$ & $0.356$ & $\underline{0.298}$ & $\underline{0.480}$ & $\underline{0.250}$ & $0.420$ & $2.432$ & $0.370$ & $35.10$ & $168.9$ & $30.35$ & $1.029$ & $3.502$ & $0.967$ \\
        CityU-02-F07   & $\textbf{0.018}$ & $\textbf{0.177}$ & $\textbf{0.010}$ & $0.078$ & $0.496$ & $0.071$ & $\underline{0.070}$ & $\underline{0.446}$ & $\underline{0.064}$ & $0.263$ & $0.872$ & $0.099$ & $0.215$ & $1.808$ & $0.182$ & $0.206$ & $1.115$ & $0.195$ \\
        CityU-02-F08   & $\textbf{0.011}$ & $\textbf{0.295}$ & $\textbf{0.024}$ & $0.100$ & $0.660$ & $0.098$ & $\underline{0.057}$ & $\underline{0.640}$ & $\underline{0.053}$ & $0.094$ & $0.702$ & $0.077$ & $21.94$ & $61.81$ & $17.24$ & $4.628$ & $6.862$ & $4.399$ \\
        CityU-02-F10   & $\textbf{0.009}$ & $\textbf{0.121}$ & $\textbf{0.009}$ & $0.095$ & $0.424$ & $0.090$ & $\underline{0.041}$ & $\underline{0.285}$ & $\underline{0.035}$ & $0.078$ & $0.364$ & $0.066$ & $15.81$ & $67.47$ & $13.56$ & $0.171$ & $0.775$ & $0.170$ \\
        CityU-02-F12   & $\textbf{0.008}$ & $\textbf{0.177}$ & $\textbf{0.008}$ & $0.107$ & $0.418$ & $0.100$ & $0.075$ & $0.376$ & $0.069$ & $0.198$ & $0.487$ & $0.117$ & $18.73$ & $100.5$ & $19.36$ & $0.282$ & $1.422$ & $0.272$ \\
        \bottomrule[1.2pt]
    \end{tabular*}
    \setlength{\tabcolsep}{6pt}
\end{table*}

\subsubsection{Benchmark:}
\label{sec:sim-benchmark}
Leveraging the BIM-robot simulation platform, we designed a comprehensive simulation benchmark that incorporates a variety of as-designed BIM models to test localization performance across different levels of difficulty, scale, and {consistency}. As shown in Figure~\ref{fig:sim-dataset}, the benchmark includes five BIM model types: 
(1) \textbf{CaseStudy}: Six simple BIM models from the ISPRS Indoor Modeling Benchmark~\citep{Khoshelham2021ISPRS}, with coverage areas ranging from {532.11}~{$\text{m}^2$} to {727.43}~{$\text{m}^2$};
(2) \textbf{CityU-01}: BIM models covering floors 06–12 of the City University of Hong Kong~\citep{Zhang2024AIC}, with a coverage area of ${80}~{\text{m}} \times {50}~{\text{m}}$;
(3) \textbf{Office-01}: A single-floor residential building model covering {3500}~{$\text{m}^2$};
(4) \textbf{Office-02}: A single-floor model of an under-construction hospital covering over {7000}~{$\text{m}^2$};
(5) \textbf{CityU-02}: A variant of CityU-01 where decorative curtain walls were removed to simulate BIM-reality discrepancies arising from unfinished structures.
During experiments, the partially removed BIM models were used in the Gazebo simulator to generate data, while the original, complete BIM models were provided to all algorithms as a reference. Data sequences are organized by building floor, with 20 sequences of varying path lengths generated: CaseStudy (7 sequences, total {885.49}{\text{m}}), CityU-01 (7 sequences, ~{7450.0}{\text{m}}), Office-01 (1 sequence, {1181.04}{\text{m}}), Office-02 (1 sequence, {670.65}{\text{m}}), and CityU-02 (4 sequences, total {4205.41}{\text{m}}). {In the aforementioned data groups, the sensor observations in CaseStudy, Office-01, Office-02, and CityU-01 are fully BIM-consistent, while those in CityU-02 are BIM-discrepant.}

\subsubsection{Baselines:}
Six state-of-the-art pure LiDAR-based methods are included for comparison: (1) offline Bundle Adjustment (BA) methods BALM2~\citep{Liu2023TRO} and HBA~\citep{Liu2023RAL}; (2) An online pose graph optimization method using global place descriptors: Scan Context~\citep{Kim2022TRO} integrated with pose graphs; (3) a map-based localization approach: HDL-Localization~\citep{Kenji2019IJARS}; (4) the DLO algorithm~\citep{Chen2022RAL}, which serves as the front-end module for BIM-Loc and is included as a standalone baseline; (5) an offline point cloud-to-BIM comparison method BIM-Reg~\citep{Zhang2024AIC} as a baseline for discrepancy detection evaluation. {Note that HDL-Localization is used solely in its \textit{local} prior-map-based pose correction mode, with global initialization disabled. The initial pose is provided in advance to ensure a fair comparison of prior-based pose optimization among all algorithms.}

\subsubsection{Metrics:}
Localization performance is evaluated using the following metrics: Absolute Trajectory Error (ATE), Mean Map Entropy (MME), RMSE of scan-to-BIM distances, and RMSE of Wasserstein Distance (WD) errors. For discrepancy detection, precision, recall, and F1-score are employed.

\begin{figure}[t]
    \centering
    \includegraphics[width=0.9\linewidth]{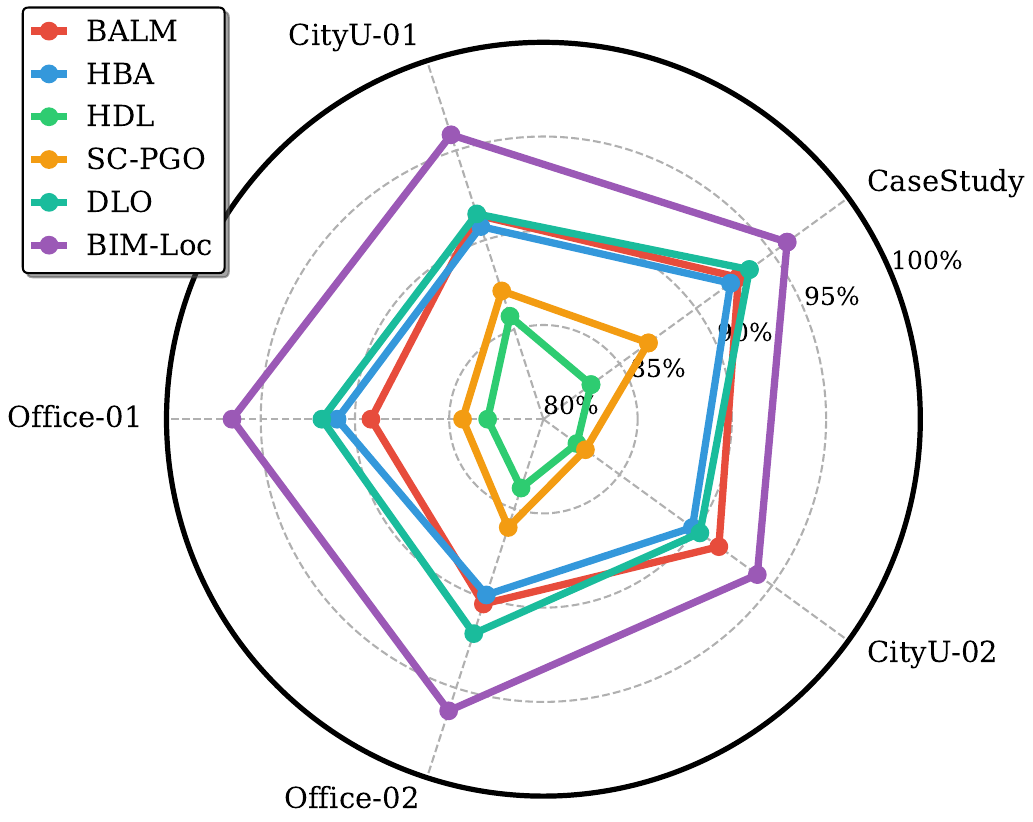}
    \caption{
    MME ratios for evaluated methods across data sequences in the simulation benchmark. Performance is grouped into three tiers, with BIM-Loc consistently ranking first.
    }
    \label{fig:sim-mme}
\end{figure}

\begin{figure*}[t]
    \centering
    \includegraphics[width=1.0\linewidth]{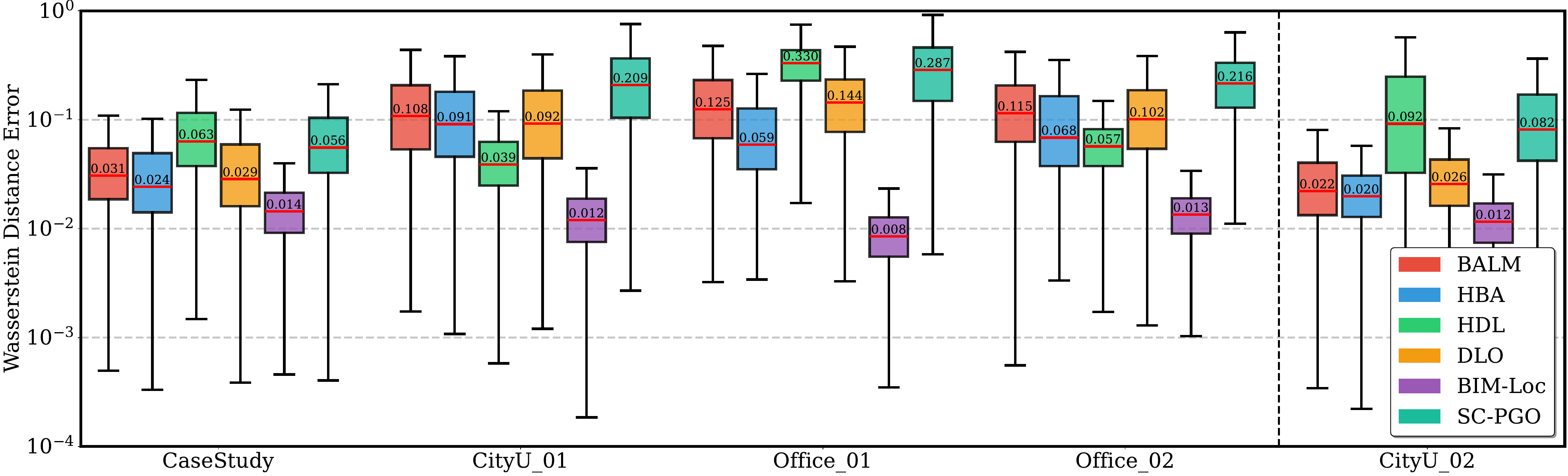}
    \caption{
Wasserstein distance errors for evaluated algorithms in the simulation benchmark. Box plots show median WD values (logarithmic $y$-axis). Results for unmodified BIM models (left of dashed line) and synthetic BIM models (right) are grouped separately.
    }
    \label{fig:sim-awd}
\end{figure*}

\begin{figure*}[t]
    \centering
    \includegraphics[width=1.0\linewidth]{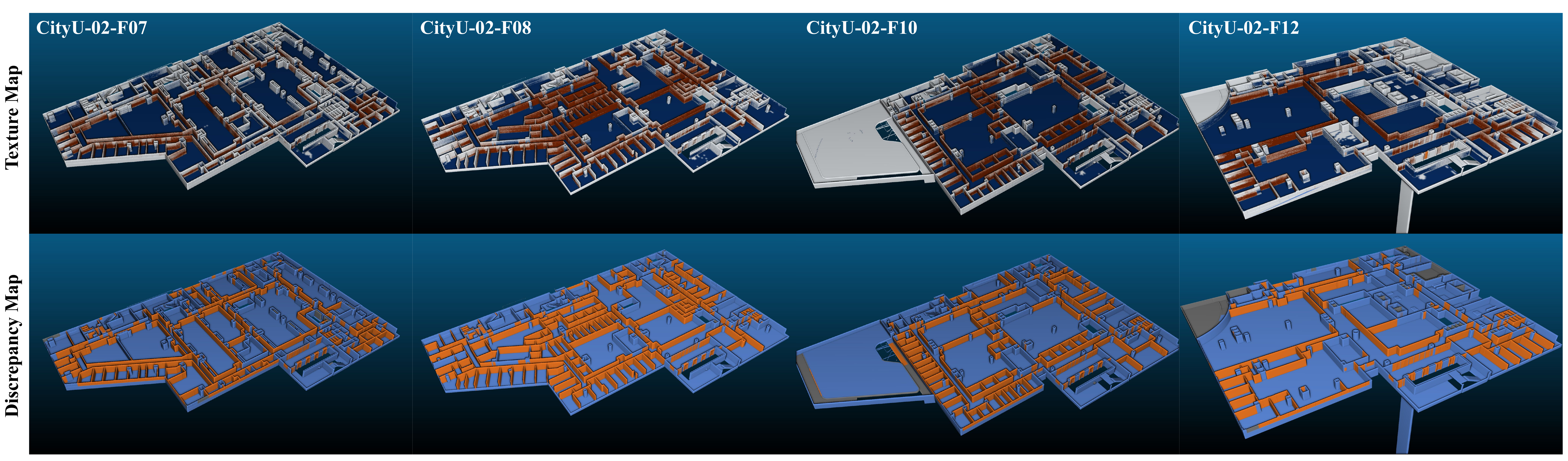}
    \caption{
    Discrepancy detection results for CityU-02 data group. Top row: texture maps for Floors 07, 08, 10, and 12. Bottom row: corresponding discrepancy maps showing built (\textcolor{blue}{blue}) and unbuilt (\textcolor{orange}{orange}) structures.
    }
    \label{fig:sim-discrepancy}
\end{figure*}

\subsubsection{Result Analysis:}
The simulation benchmark results are presented below according to the evaluation metrics. First, as shown in Table~\ref{tab:ape-rmse-results}, our BIM-Loc method achieves the best performance across all 20 sequences, {with the best RMSE values of 0.007 m for translation, 0.115 degrees for rotation, and 0.007 m for scan-to-BIM distances.} These lowest ATE RMSE values confirm BIM-Loc's superior localization accuracy, while the minimal scan-to-BIM distances indicate optimal alignment between scans and the BIM model. Among the other evaluated algorithms, both the online map-based method (HDL) and offline BA-based methods outperform the DLO front-end, demonstrating that pose graph optimization and map priors can enhance raw odometry results. However, these methods cannot fully eliminate accumulated drift errors due to their reliance on voxelized point clouds for data association. HBA ranks second best in the CaseStudy, Office-02, and CityU-02 sequences, while HDL performs second best in CityU-01. This variation stems from the voxel-based association mechanisms in HBA and BALM2, which are misled by the narrow double sides of thin walls, as noted by~\cite{Zhou2021RAL} and visualized in Figure~\ref{fig:thin-wall}. The complex layout of CityU-01, with its numerous thin decorative walls, exacerbates this issue. In contrast, BIM-Loc's multi-hit ray casting module clearly distinguishes both sides of thin walls, enabling their effective incorporation into the optimization. Consequently, BIM-Loc demonstrates better localization performance and greater robustness across all building structure experiments. {Additionally, due to the fact that the experiments in CaseStudy and CityU-01 are conducted in BIM-consistent regions, i.e. without BIM-reality discrepancies. It can prove that the performance gains in above settings are mostly owning to optimizations with BIM priors.}

In Figure~\ref{fig:sim-mme}, the MME ratios are divided into three performance tiers. BIM-Loc alone occupies the top tier with a ratio of $95.73\pm{0.99}\%$, significantly outperforming the DLO odometry ($91.79\pm{1.17}\%$) in the second tier. HBA and BALM2 also fall into the second tier ($90.71\pm{1.03}\%$ and $91.03\pm{1.37}\%$, respectively). The substantial gap between the first and second tiers underscores the impact of incorrect data association in HBA and BALM2, reducing their performance to the level of DLO odometry. SC-PGO and HDL belong to the third tier. BIM-Loc's superior MME ratios reflect excellent local spatial consistency in the aggregated map and, according to~\cite{Kornilova2021ECMR}, indicate the lowest relative pose errors among all methods, confirming its overall advantage.

Figure~\ref{fig:sim-awd} presents box plots of Wasserstein Distance (WD) errors, evaluating global consistency between the aggregated and BIM maps. BIM-Loc achieves the lowest WD errors by effectively integrating BIM models. As with the ATE and MME evaluations, the complex layouts of CityU-01 degrade the performance of offline BA-based methods (HBA, BALM2) compared to CityU-02. Conversely, HDL benefits from map priors and performs better in CityU-01, confirming that map priors can enhance indoor localization accuracy. Overall, BIM-Loc maintains strong performance across all sequences by deeply integrating BIM models into a pose graph optimization framework, effectively bounding localization errors and ensuring tight consistency within the BIM coordinate system.

\begin{figure*}[t]
    \centering
    \includegraphics[width=1.0\linewidth]{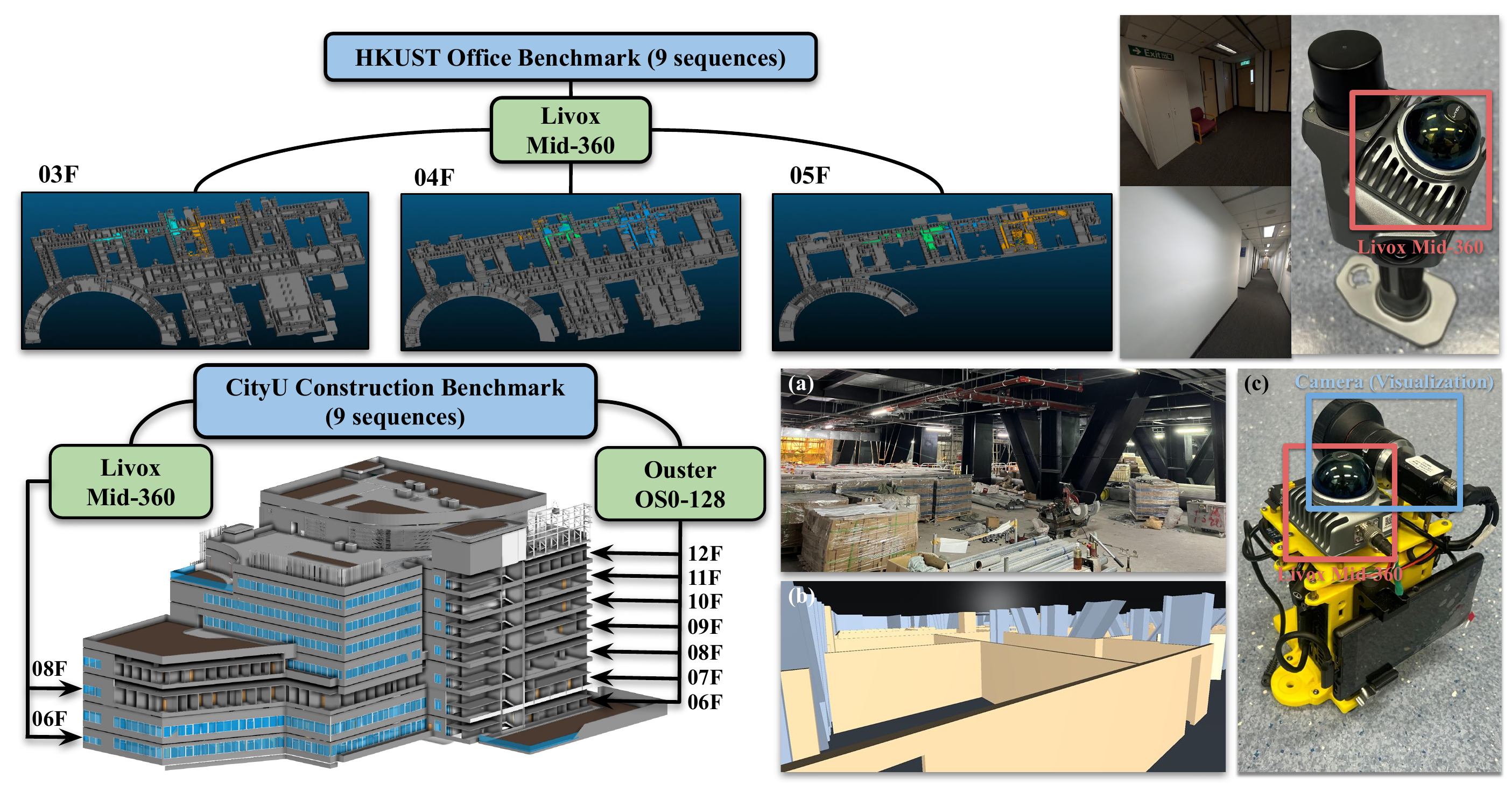}
    \caption{
    Real-world evaluation benchmarks representing different building lifecycle phases. The HKUST office benchmark assesses localization in a completed building using a handheld Livox Mid-360 sensor with built-in IMU. The CityU construction benchmark evaluates performance in an active construction environment with significant BIM-reality discrepancies, using both Ouster OS0-128 and Livox Mid-360 sensors to enhance evaluation diversity.
    }
    \label{fig:real-benchmark}
\end{figure*}

\begin{table}[t]
  \centering
  \caption{Precision, recall, and F1-score values of discrepancy detection performance in four CityU-02 sequences in simulation benchmark.}
  \label{tab:sim-discrepancy}
  \renewcommand{\arraystretch}{1.2}
  \scriptsize
  \begin{tabular*}{240pt}{@{\extracolsep\fill}l>{\raggedright\arraybackslash}p{2.0cm}ccc@{\extracolsep\fill}}
    \toprule[1.2pt] 
    \textbf{Algorithm} & \textbf{Sequence} & \textbf{Recall} & \textbf{Precision} & \textbf{F1-Score} \\ 
    \midrule[0.8pt] 
    \multirow{5}{*}{BIM-Loc} 
    & CityU-02-F07 & 0.981 & 0.877 & 0.926 \\ 
    & CityU-02-F08 & 1.000 & 0.873 & 0.932 \\ 
    & CityU-02-F10 & 0.992 & 0.852 & 0.917 \\ 
    & CityU-02-F12 & 0.915 & 0.844 & 0.878 \\ 
    \cmidrule{2-5}
    & Average & 0.972 & 0.861 & 0.913 \\
    \midrule[1.2pt]
    \multirow{5}{*}{BIM-Reg} 
    & CityU-02-F07 & 0.819 & 0.546 & 0.655 \\
    & CityU-02-F08 & 0.911 & 0.661 & 0.766 \\
    & CityU-02-F10 & 0.789 & 0.647 & 0.711 \\
    & CityU-02-F12 & 0.718 & 0.624 & 0.668 \\
    \cmidrule{2-5}
    & Average & 0.809 & 0.619 & 0.700 \\
    \bottomrule[1.2pt]
  \end{tabular*}
\end{table}

Table~\ref{tab:sim-discrepancy} evaluates the discrepancy detection performance on four CityU-02 sequences. BIM-Loc substantially outperforms the BIM-Reg algorithm, with an average F1-score of $0.913$ versus $0.700$. Notably, BIM-Loc achieves a higher recall ($0.972$) than precision ($0.861$), indicating that its discrepancy detection module (Section~\ref{sec:discrepancy-identification}) prioritizes identifying all potential discrepancies to support localization tasks. This conservative approach may occasionally misclassify a finished structure as incomplete, but ensures that only verified completed structures contribute to optimization, safeguarding localization accuracy within the BIM coordinate system.

Such conservative estimates have minimal negative impact in practice, as shown by the visualization of discrepancy detection results for the four CityU-02 sequences in Figure~\ref{fig:sim-discrepancy}. The wrapped texture maps (first row) demonstrate the performance of BIM surface mapping via Bayesian kernelized inference. As described in Section~\ref{sec:multi-hit}, hit and miss statuses from multi-hit ray casting are recorded on the texture maps. The pixel-level details in these texture images clearly distinguish decorative walls from surrounding structures, corroborating the high recall values in Table~\ref{tab:sim-discrepancy}. Using the hierarchical probabilistic model (Section~\ref{sec:hierarchical-model}), discrepancy predictions for each structure are generated and displayed as discrepancy maps (second row). These maps use a three-color scheme: blue for finished structures, orange for unfinished structures, and gray for unknown regions due to insufficient observation.

\subsection{Real-World Application}
\label{sec:real-test}
To complement the simulation benchmark, we conducted a series of real-world experiments. A key distinction from simulation is that real-world scans often contain significant occlusions from objects not represented in the BIM models. These occlusions can mislead algorithms, leading to incorrect data association and degraded localization performance. Additionally, similar to the CityU-02 simulation scenario, unfinished building structures further challenge localization accuracy. To fully evaluate the impact of BIM-reality inconsistencies throughout the building lifecycle, we selected two distinct environmental settings with corresponding BIM models: one representing the construction phase and another representing the usage phase.

The first set of experiments utilizes the SLABIM dataset~\citep{Huang2025ICRA}, which includes BIM models of a multi-floor office building at the Hong Kong University of Science and Technology (HKUST). Data were collected using a handheld sensor suite featuring a Livox Mid-360 LiDAR sensor with a built-in IMU. This dataset captures typical office environments including corridors, rooms, lobbies, and lounges. As shown in the upper part of Figure~\ref{fig:real-benchmark}, discrepancies between the BIM and reality primarily stem from interior decorations, furnishings, and other office elements not modeled in the BIM, which contains only structural components (walls, floors, columns, doors). This setup provides a realistic testbed for evaluating localization adaptability in operational settings and assessing potential for service robot applications.

The second set of experiments focuses on the construction phase, using the same CityU BIM models from the CityU-01 and CityU-02 simulation benchmarks. Unlike the synthetic data, these sequences were recorded directly at active construction sites using two different LiDAR sensors (Ouster OS0-128 and Livox Mid-360) to enhance data diversity. Each data sequence includes LiDAR scans and synchronized IMU measurements. As illustrated in the bottom part of Figure~\ref{fig:real-benchmark}, the CityU construction sites exhibit significant BIM-reality discrepancies arising from unfinished structures, construction materials, and temporary equipment (e.g., ladders, pipes, trolleys). This scenario validates localization performance under challenging conditions and demonstrates the method's potential for construction inspection tasks.

\subsubsection{Benchmark:}
\label{sec:real-benchmark}
Using the available BIM models and sensor data, we established two distinct real-world benchmarks: an office benchmark (HKUST) and a construction benchmark (CityU). The data organization for both is illustrated in Figure~\ref{fig:real-benchmark}. For the office benchmark, data from Floors 03 to 05 of the SLABIM dataset were selected. Each floor includes three sequences corresponding to different regions (01, 02, and 03), resulting in a total of 9 sequences with a combined path length of 1920~m. For the construction benchmark, data were organized by floor and sensor type. Seven sequences were collected using a handheld Ouster OS0-128 sensor suite, covering construction sites from Floor 06 to Floor 12. Additionally, data sequences from Floor 06 and 08 were recorded using a Livox Mid-360 sensor. A camera was included in the sensor suite for visualization purposes only. In total, 9 data sequences with a combined path length exceeding 3.5~km were used for evaluation.

\begin{table}[t]
    \caption{
    RMSE of ATE values for the HKUST office benchmark. Each cell reports RMSE values for translation error [m] $\mid$ rotation error [degree]. Lower values indicate better performance.
    }
    \label{tab:real-slabim-ate}
    \centering
    \renewcommand{\arraystretch}{1.25}
    \scriptsize
    \begin{tabular*}{240pt}{@{\extracolsep\fill}p{1.2cm}c|cc@{\extracolsep\fill}}
        \toprule[1.2pt]
        \footnotesize\textbf{Sequence} & \footnotesize\textbf{BIM-Loc} & \footnotesize\textbf{Fast-Loc} & \footnotesize\textbf{PALoc} \\
        \midrule
        3F-Region1   & $\textbf{0.147}|\textbf{0.748}$ & $10.75|40.61$ & $\underline{0.359}|\underline{4.036}$ \\
        3F-Region2   & $\textbf{0.144}|\textbf{0.565}$ & $1.271|\underline{9.468}$  & $\underline{1.121}|12.99$ \\
        3F-Region3   & $\textbf{0.055}|\textbf{0.470}$  & $3.958|115.4$ & $\underline{0.170}|\underline{3.144}$  \\
        \midrule
        4F-Region1   & $\textbf{0.098}|\textbf{0.668}$ & $17.46|105.7$ & $\underline{0.353}|\underline{2.581}$ \\
        4F-Region2   & $\textbf{0.053}|\textbf{0.558}$ & $21.13|123.9$ & $\underline{0.374}|\underline{2.510}$  \\
        4F-Region3   & $\textbf{0.033}|\textbf{0.397}$ & $1.624|16.48$  & $\underline{0.231}|\underline{4.181}$ \\
        \midrule
        5F-Region1   & $\textbf{0.076}|\textbf{2.911}$ & $14.04|75.21$ & $\underline{0.455}|\underline{4.046}$ \\
        5F-Region2   & $\textbf{0.043}|\textbf{0.402}$ & $27.36|138.9$ & $\underline{0.126}|\underline{2.952}$ \\
        5F-Region3   & $\textbf{0.055}|\textbf{0.632}$ & $19.34|36.50$ & $\underline{0.836}|\underline{4.286}$ \\
        \bottomrule[1.2pt]
    \end{tabular*}
\end{table}

\begin{table}[t]
  \centering
  \caption{RMSE of scan-to-BIM distance errors [m] on HKUST Office benchmark. Lower values indicate better performance.}
  \label{tab:scan-to-bim-rmse-hkust}
  \renewcommand{\arraystretch}{1.25}
  \scriptsize
  \begin{tabular*}{240pt}{@{\extracolsep\fill}l*{3}{>{\centering\arraybackslash}p{1.85cm}}@{\extracolsep\fill}}
    \toprule[1.2pt]
    \textbf{Sequence} & \textbf{BIM-Loc} & \textbf{Fast-Loc} & \textbf{PALoc} \\
    \midrule[0.8pt]
    3F-Region1 & \textbf{0.052} & \underline{0.067} & 0.077 \\
    3F-Region2 & \textbf{0.037} & \underline{0.062} & 0.065 \\
    3F-Region3 & \textbf{0.044} & 0.067 & \underline{0.061} \\
    \cmidrule{1-4}
    4F-Region1 & \textbf{0.044} & 0.102 & \underline{0.073} \\
    4F-Region2 & \textbf{0.047} & \underline{0.063} & 0.068 \\
    4F-Region3 & \textbf{0.045} & 0.068 & \underline{0.061} \\
    \cmidrule{1-4}
    5F-Region1 & \textbf{0.054} & 0.085 & \underline{0.067} \\
    5F-Region2 & \textbf{0.039} & \underline{0.065} & 0.066 \\
    5F-Region3 & \textbf{0.045} & 0.082 & \underline{0.081} \\
    \bottomrule[1.2pt]
  \end{tabular*}
\end{table}

\subsubsection{Baselines:}
\label{sec:real-baseline}
Two state-of-the-art map-based algorithms utilizing both LiDAR and IMU data were evaluated in these real-world scenarios: (1) Fast-LIO-Localization (Fast-Loc)~\citep{fast-loc}, which extends the original Fast-LIO with a global alignment operation between map priors and the locally accumulated map; (2) the recently published PALoc~\citep{Hu2024TMech}, a state-of-the-art map-based localization approach. {Additionally, CAD-Mesher~\citep{Jia2025TMM} and LIO-BIM~\citep{Stuhrenberg2025AEI} are included to evaluate localization performance under different map representations.} BIM-Loc utilizes the raw BIM data directly, while for the other two methods, point cloud maps were generated by sampling points from the BIM model surfaces. All algorithms were evaluated on the same data sequences. Note that point clouds reconstructed from real-world observations were not available for any of these methods.

\subsubsection{Metrics:}
\label{sec:real-metrics}

\begin{table}[t]
  \centering
  \caption{{RMSE of scan-to-BIM distance errors [m] on CityU Construction benchmark. Lower values indicate better performance.}}
  \label{tab:scan-to-bim-rmse-cityu}
  \renewcommand{\arraystretch}{1.25}
  \scriptsize
  \begin{tabular*}{240pt}{@{\extracolsep\fill}l*{5}{>{\centering\arraybackslash}p{0.9cm}}@{\extracolsep\fill}}
    \toprule[1.2pt]
    \textbf{Sequence} & \textbf{BIM-Loc} & \textbf{Fast-Loc} & \textbf{PALoc} & \textbf{LIO-BIM} & \textbf{CAD-Mesher} \\
    \midrule[0.8pt]
    Floor-06 (Mid-360) & \textbf{0.036} & \underline{0.080} & 0.087 & -- & -- \\
    Floor-08 (Mid-360) & \textbf{0.038} & 0.070 & \underline{0.057} & -- & -- \\
    \cmidrule{1-6}
    Floor-06 (Ouster) & \textbf{0.045} & 0.078 & 0.073 & \underline{0.061} & 0.062 \\
    Floor-07 (Ouster) & \textbf{0.046} & 0.076 & 0.068 & \underline{0.059} & 0.064 \\
    Floor-08 (Ouster) & \textbf{0.041} & 0.072 & 0.064 & \underline{0.056} & 0.065 \\
    Floor-09 (Ouster) & \textbf{0.044} & 0.073 & 0.060 & \underline{0.059} & 0.065 \\
    Floor-10 (Ouster) & \textbf{0.042} & 0.074 & 0.064 & \underline{0.056} & 0.062 \\
    Floor-11 (Ouster) & \textbf{0.044} & 0.076 & 0.067 & \underline{0.059} & 0.062 \\
    Floor-12 (Ouster) & \textbf{0.042} & 0.071 & 0.062 & \underline{0.060} & \underline{0.060} \\
    \bottomrule[1.2pt]
  \end{tabular*}
\end{table}

For the office benchmark, which possesses ground truth trajectories, localization performance was evaluated using ATE, MME, scan-to-BIM distance errors, and Wasserstein distance errors. For the construction benchmark, where ground truth trajectories were unavailable, performance was assessed using MME, scan-to-BIM distance errors, and Wasserstein distance errors. In addition, the discrepancy detection results of BIM-Loc were visualized, then its localization performance was further validated qualitatively by comparing images from the actual scenes with corresponding renderings from the BIM model based on the estimated poses.

\begin{figure*}[t]
    \centering
    \begin{subfigure}[b]{0.48\linewidth} 
        \centering
        \includegraphics[width=0.95\linewidth]{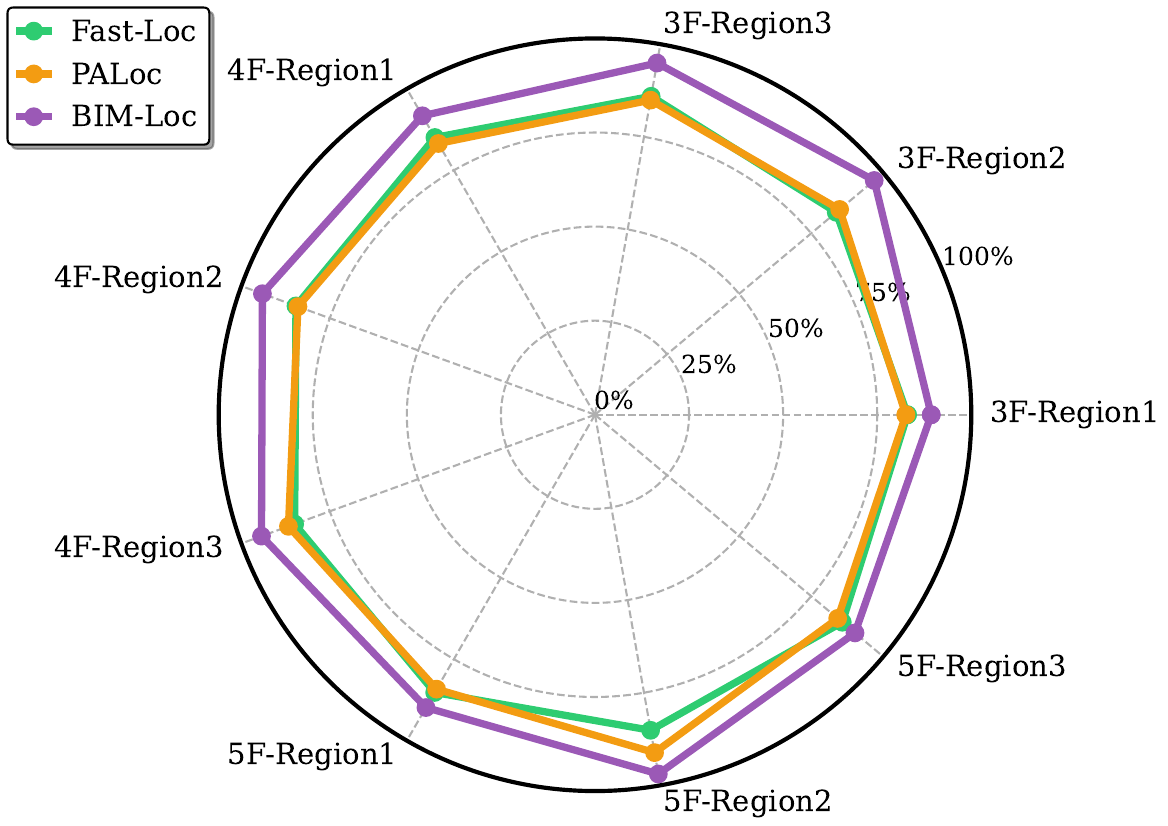}
        \caption{MME ratios for HKUST office benchmark}
        \label{fig:slabim-mme}
    \end{subfigure}
    \hfill
    \begin{subfigure}[b]{0.48\linewidth}
        \centering
        \includegraphics[width=1.0\linewidth]{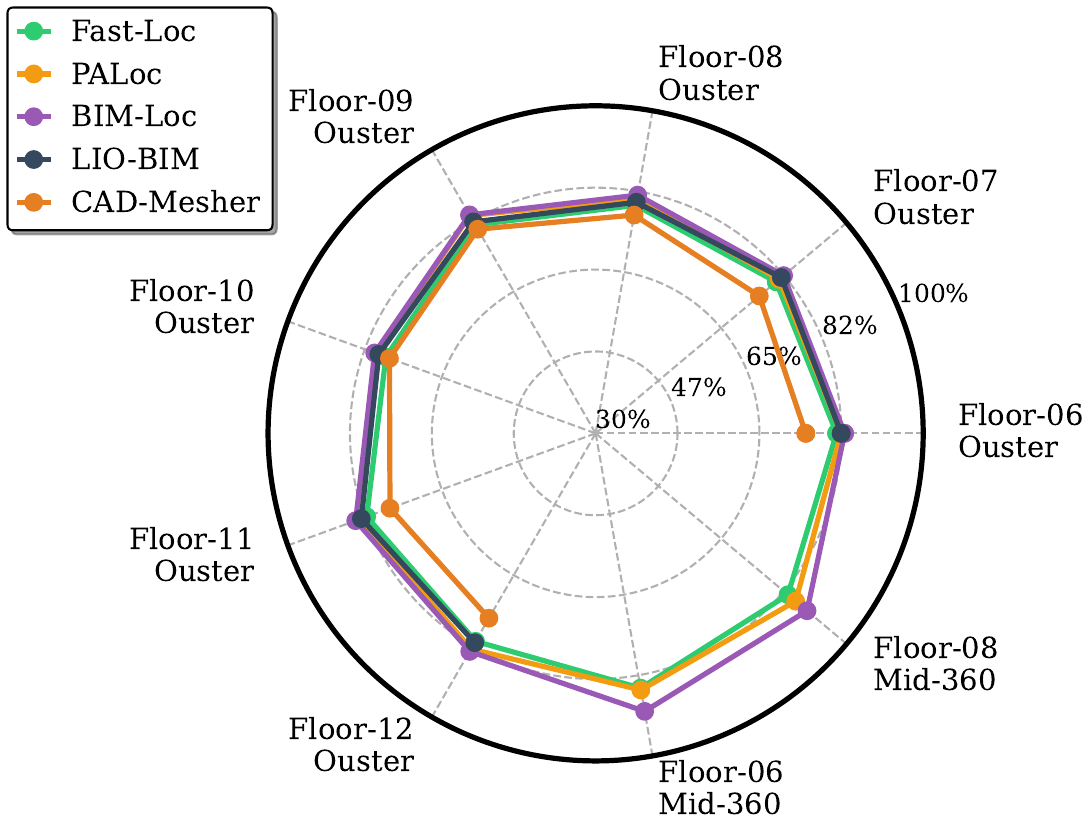} 
        \caption{{MME ratios for CityU construction benchmark.}}
        \label{fig:construction-mme}
    \end{subfigure}
    \caption{
    MME ratios for evaluated methods in two real-world benchmarks. BIM-Loc achieves the highest ratios, with a substantial performance gap in the HKUST office benchmark and a smaller but consistent lead in CityU construction benchmark.
    }
    \label{fig:real-mme} 
\end{figure*}

\begin{figure*}[ht]
    \centering
    \includegraphics[width=1.0\linewidth]{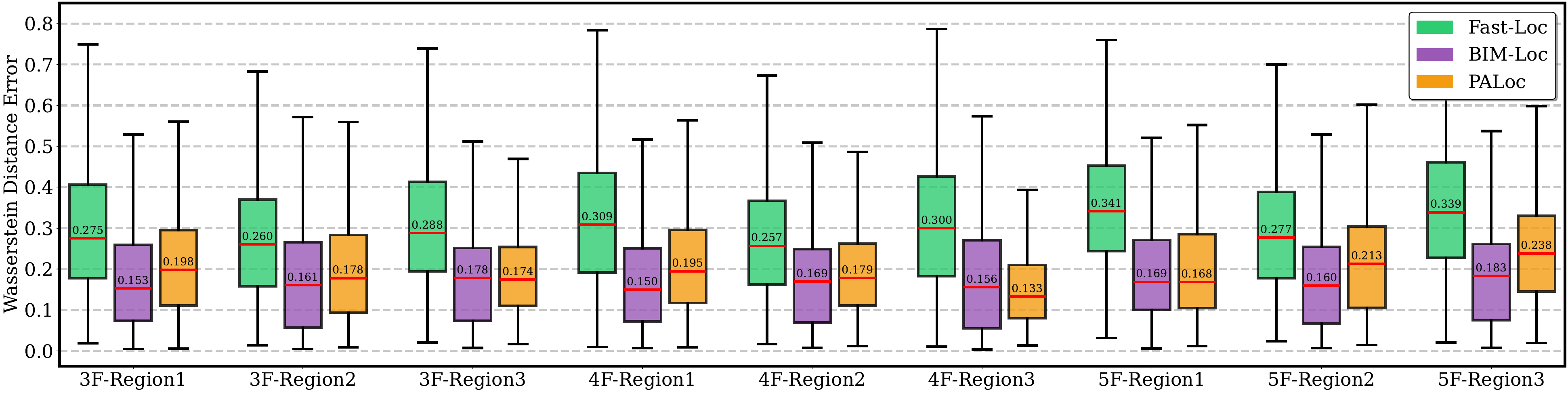}
    \caption{
    Wasserstein distance errors for evaluated methods in the HKUST office benchmark. Red lines indicate median WD errors within each box plot.
    }
    \label{fig:slabim-awd}
\end{figure*}

\begin{figure*}[ht]
    \centering
    \includegraphics[width=1.0\linewidth]{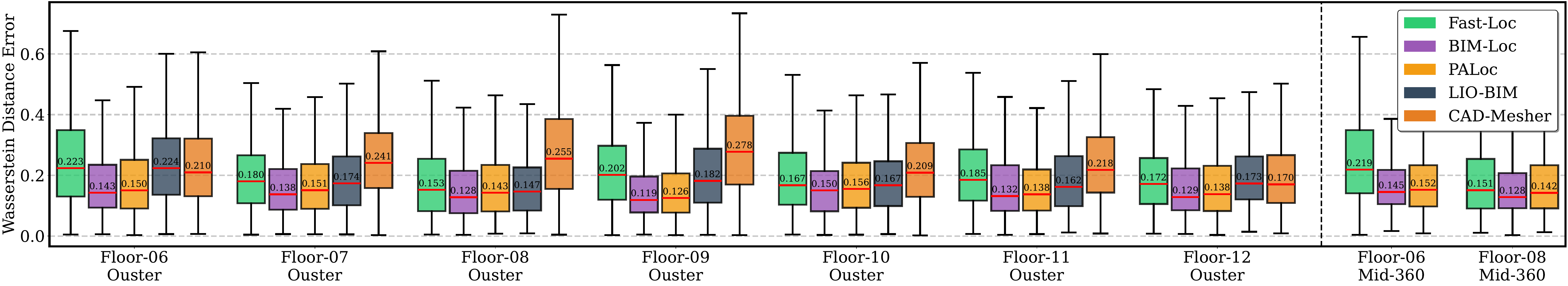}
    \caption{
    {Wasserstein distance errors for evaluated methods in the CityU construction benchmark. Results left of the dashed line correspond to Ouster OS0-128 sensor data; right side corresponds to Livox Mid-360 sensor data. Red lines indicate median WD errors.}
    }
    \label{fig:cityu-awd}
\end{figure*}

\begin{figure*}[t]
    \centering
    \includegraphics[width=1.0\linewidth]{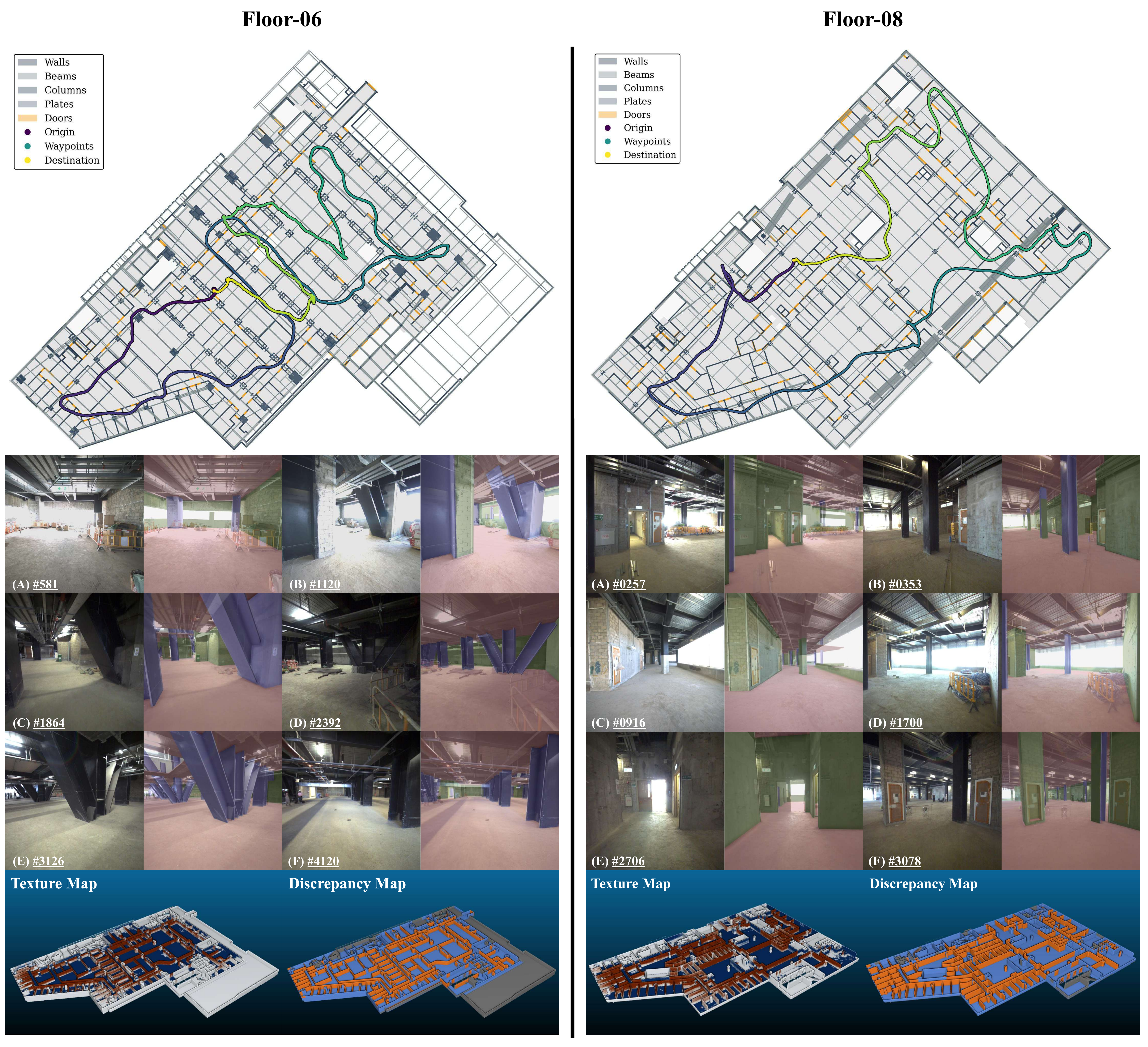}
    \caption{
    Localization and discrepancy detection results of BIM-Loc in the CityU construction benchmark. Top row: estimated trajectories in BIM coordinates for Floors 06 and 08. Middle rows: real construction site images (A, C, E) and corresponding BIM-rendered overlays (B, D, F). Bottom row: texture maps and discrepancy maps generated during localization.
    }
    \label{fig:cityu-vis}
\end{figure*}

\subsubsection{Result Analysis:}
\label{sec:real-analysis}
The experimental results from the two benchmarks are organized and presented below according to the evaluation metrics. We first describe the results from the office benchmark, followed by those from the construction benchmark.

\noindent\textbf{Office Benchmark} provides ground truth trajectories from the SLABIM dataset. In terms of ATE {in Table~\ref{tab:real-slabim-ate}, BIM-Loc outperforms other map-based localization methods, having the RMSE values of 0.147~m in translation and $2.911^{\circ}$ in rotation in the worst case. In contrast, Fast-Loc exhibits severe drift, with the worst RMSE values as high as {27.36}{m} (translation) and ${138.9}^{\circ}$ (rotation). PALoc ranks second, showing relatively stable performance with the worst translation and rotation RMSE values of {1.121}{m} and $12.99^{\circ}$, respectively.} While these values are notably higher than BIM-Loc's worst results, they are much lower than those of Fast-Loc. This performance gap indicates that PALoc heavily relies on map priors from actual scenes to achieve high localization accuracy, and drift errors cannot be fully mitigated using BIM models alone due to BIM-reality discrepancies. In contrast, BIM-Loc demonstrates greater robustness by deeply integrating BIM models.

As shown in Figure~\ref{fig:slabim-mme}, a significant gap in MME ratios exists between BIM-Loc ($93.13\pm{2.94}\%$) and the other two methods (Fast-Loc: $84.82\pm{0.94}\%$, PALoc: $85.10\pm{2.55}\%$). This gap is consistent across all test sequences. Although Fast-Loc and PALoc alternate in securing the second place across individual sequences, their average MME ratios are nearly identical (84.82\% vs. 85.10\%).

The scan-to-BIM distance evaluation (Table~\ref{tab:scan-to-bim-rmse-hkust}) further confirms BIM-Loc's superiority, with an average RMSE of {0.046}{m}, outperforming Fast-Loc ({0.074}{m}) and PALoc ({0.069}{m}). BIM-Loc consistently achieves the lowest RMSE values in every sequence, ranging from {0.037}{m} (3F-Region2) to {0.054}{m} (5F-Region1). This underscores its ability to maintain strong consistency between LiDAR scans and the BIM model. In comparison, Fast-Loc and PALoc exhibit higher RMSE values, indicating weaker BIM-observation consistency. Note that because the scan-to-BIM distance is truncated, Fast-Loc's results are only reliable when drift errors are minimal.

Similar trends are observed in the WD errors (Figure~\ref{fig:slabim-awd}). BIM-Loc consistently shows the lowest first quartile (Q1) values and maintains smaller error magnitudes, reflecting its stable performance when using BIM as a reference. The median error values (represented by the red line in the box plots) also highlight BIM-Loc's advantage across regions. Additionally, BIM-Loc's error spread is narrower in most sequences, with no extreme high-error outliers that may appear in Fast-Loc or PALoc, further proving its resilience to environmental variations across different floors and regions in complex office scenarios.

\noindent\textbf{Construction benchmark} evaluates performance in construction sites, where ground truth trajectories are unavailable. Thus, localization accuracy is assessed using MME ratios, scan-to-BIM distances, and WD errors. Compared to the simulation and office benchmarks, BIM-Loc's MME ratios show a slight decrease, reflecting the challenging nature of construction environments. Nonetheless, BIM-Loc still outperforms Fast-Loc and PALoc across all floors and sensor types.
As illustrated in Figure~\ref{fig:construction-mme}, with the Ouster OS0-128 sensor, BIM-Loc's MME ratios range from 80.25\% to 84.57\%, demonstrating high and stable performance. PALoc ranks second, with ratios between 80.83\% and 83.80\%, always lower than BIM-Loc. The gap widens to 0.72\%–2.54\% when comparing the same sequences. With the Livox Mid-360 sensor, BIM-Loc achieves exceptional MME ratios of 90.32\% (Floor-06) and 88.98\% (Floor-08), exceeding its performance with the Ouster sensor (max. 84.57\%). This highlights BIM-Loc's compatibility with the Mid-360 sensor, likely due to its high-ranging accuracy. The other methods struggle to match this performance: PALoc reaches 85.69\% (Floor-06) and 85.80\% (Floor-08), while Fast-Loc achieves only 85.28\% (Floor-06) and 83.65\% (Floor-08). The gaps between BIM-Loc and the others expand to 4.63\%–5.33\% (vs. PALoc) and 3.64\%–6.67\% (vs. Fast-Loc), underscoring BIM-Loc's adaptability to diverse sensors.

BIM-Loc's advantages are further validated by scan-to-BIM distances and WD errors. In Table~\ref{tab:scan-to-bim-rmse-cityu}, BIM-Loc achieves the lowest RMSE values across all sequences. For Livox-captured data, it records {0.036}{m} (Floor-06) and {0.038}{m} (Floor-08), outperforming both baselines. Even PALoc's best result ({0.057}{m} on Floor-08) is {0.019}{m} higher. For Ouster-captured sequences (Floors 06--12), BIM-Loc maintains consistently low RMSE values ({0.041} to {0.046}{m}), significantly lower than Fast-Loc ({0.071} to {0.078}{m}) and PALoc ({0.060} to {0.073}{m}). {The two structured-map baselines, LIO-BIM and CAD-Mesher, evaluated on the same Ouster sequences, also fall short of BIM-Loc: LIO-BIM attains {0.056}to{0.061}{m} (second-best per sequence in the table) and CAD-Mesher {0.060} to {0.065}{m}, both clearly higher than BIM-Loc's {0.041} to {0.046}{m}.} These consistent gaps demonstrate BIM-Loc's ability to ensure precise scan-to-BIM alignment in construction scenarios, which is critical for progress monitoring and quality inspection.

WD errors (Figure~\ref{fig:cityu-awd}) show that BIM-Loc maintains the best overall performance in terms of global consistency between aggregated maps and BIM models. For Ouster sequences, its errors cluster tightly between 0.119 and 0.150 (e.g., 0.119 on Floor-09 vs. PALoc's 0.126). For Livox sequences, it retains an edge with values of 0.145 (Floor-06) and 0.128 (Floor-08). The narrow data range of BIM-Loc's errors further confirms the consistency between BIM and accumulated scans.

Furthermore, localization performance is visually validated through camera imagery. Using the sensor suite from Figure~\ref{fig:real-benchmark}, we compare real-world footage from CityU construction sites with BIM-rendered images from the same camera poses. As shown in Figure~\ref{fig:cityu-vis} (rows 2–5), six image pairs (A–F) are presented for each floor. Each pair includes a real scene image (A, C, E) and a blended image overlaying the BIM rendering (B, D, F). For clarity, only structures present in the real scenes are included in the BIM renderings, with color-coding for different types (e.g., walls: olive green, floors: red, H-shape beams: purple). In Floors 06 and 08, the purple H-shape beams in the BIM renderings show perfect overlap with their real-world counterparts throughout the localization tasks.

The final row of Figure~\ref{fig:cityu-vis} illustrates the wrapped texture and discrepancy maps. Due to incomplete scan coverage, the texture maps use three colors to represent status: miss (orange), hit (blue), and unknown (white). These maps reflect data coverage during inspection. Decoration walls are accurately recorded and labeled in both texture and discrepancy maps. For example, in Floor 06, an H-shape beam embedded in a concrete column is not observed during inspection but is correctly labeled as unknown. Additionally, triangular H-shape beams behind tightly covered decoration walls are clearly distinguishable.

\section{Discussion}
\label{sec:discussion}
{In our BIM-Loc method, BIM prior is the dominant contributor: by constraining estimates to structurally plausible configurations, it yields the largest quantitative gains in accuracy and long-term robustness. Discrepancy handling provides a secondary yet essential improvement: its primary role is defensive, refining residuals arising from unmodeled variations and preventing false constraints from degrading performance. Data association operates as an algorithmic enabler: it establishes reliable BIM-to-sensor correspondences in real time without introducing computational overhead. Thus, the quantitative performance gains stem mainly from the BIM prior and discrepancy handling, while data association preserves the real-time capability that makes these gains practically attainable. A few details of BIM-Loc method will be discussed in the following sections.} 

\subsection{Computational Scalability}
\label{sec:scalability}
{In design, BIM-Loc is computationally scalable despite the seemingly intensive components (multi-hit ray casting, facet-based clustering, and hierarchical Bayesian updates). BIM-aided trajectory optimization uses iSAM2 in GTSAM for incremental inference with partial relinearization and variable reordering, so the per-step cost grows only with the locally affected region rather than the full trajectory. Second, multi-hit ray casting relies on a BVH built once offline with $O(N \log N)$ complexity in the number of BIM triangles; at runtime, the cost is dominated by the number of rays and their maximum range and is effectively decoupled from global mesh size. Third, facet-based clustering exploits a precomputed dictionary from (Geometric ID, Primitive ID) to facet IDs, allowing $O(1)$ lookups. Hierarchical Bayesian updates are also applied incrementally on this sparse factor graph, supporting deployment in large, complex buildings. For details of runtime analysis, please refer to Appendix~\ref{sec:runtime-analysis}.}

\subsection{BIM Map for Localization Tasks}
\label{sec:bim-advantages}
BIM models play a pivotal role in our BIM-Loc method. Their technical characteristics align deeply with the requirements of practical indoor localization, offering significant advantages over traditional 3D representations like point clouds and voxels.

First, the lightweight nature of BIM effectively addresses storage and computational challenges. While point clouds accurately capture spatial morphology, their massive data volumes impose high memory and computational burdens during data loading. In contrast, BIM employs structured geometric topology and parametric attribute information (e.g., for walls and beams) to compress data volume, making it more suitable for mobile or embedded devices with limited computing resources. For example, in the construction benchmark, the BIM models for Floors 06 and 08 consume approximately {12.0}{MB} and {10.8}{MB}, respectively, whereas point clouds sampled from the same surfaces require {168.4}{MB} and {152.7}{MB}.

Second, BIM provides more stable geometric benchmarks for positioning. Point clouds are susceptible to scanning noise and occlusions, leading to uneven density and feature loss that can compromise the accuracy of normal vectors and other reference features. BIM, however, is built on design specifications or reconstructed structured models, ensuring clear topological relationships and consistent surface normals and corner features. This results in a more reliable positioning reference. Furthermore, BIM's practicality extends across the entire building lifecycle. For new constructions, BIM is inherently available from planning and design phases, eliminating the need for additional modeling costs. For existing buildings, although laser scanning is required for reconstruction, the resulting BIM model supports long-term reuse, significantly reducing the storage and update overhead for spatial data in localization systems.

\subsection{Texture Map for Discrepancy Mapping}
\label{sec:texture-advantages}
We selected texture mapping as the foundation for discrepancy detection due to its advantages in representation directness, storage efficiency, and resolution adaptability, which are superior to voxel-based association methods.

For information representation, texture maps directly encode the physical coverage status of building surfaces without requiring complex data association or feature mapping. Point clouds need interpolation to form continuous surfaces, a process prone to errors, while voxel methods rely on 3D grid discretization and filling rates, which struggle to accurately capture local surface discrepancies.

In terms of storage efficiency, the difference between 2D texture maps and 3D voxel data is substantial. For a {4000}~{$\text{m}^2$} space like Floor 06 of the CityU construction benchmark, a texture map consumes approximately {5.0}{MB}, whereas a Truncated Signed Distance Function (TSDF) voxel representation requires about {823.4}{MB}. This demonstrates that texture maps significantly reduce the storage and transmission costs of discrepancy detection systems.

Additionally, texture maps support adaptive resolution {automatically}, enhancing both flexibility and accuracy. {Under this setting, the texel size (\textbf{physical area per pixel in real world}) is inversely proportional to the geometric surface area of the structural element. For \textbf{High-curvature, small-area structures} (e.g., H-shape beams), each pixel represents approximately $0.01\mathrm{m}^2$ of surface area, allowing fine-grained detail capture for narrow structural elements. For \textbf{Large planar surfaces} (e.g., floor slabs), each pixel may represent approximately $1.0\mathrm{m}^2$ of surface area, maintaining sufficient semantic information while preventing excessive memory consumption.} Large-scale flat structures can be efficiently processed with low-resolution textures (e.g., $512\times512$ pixels for a {4000}~{$\text{m}^2$} slab), while small details like beams and columns can be captured with high-resolution textures (e.g., $2048\times2048$ pixels) to preserve subtle discrepancies. This structure-oriented resolution strategy avoids the detail loss of fixed low resolutions and the computational waste of fixed high resolutions, achieving an optimal balance between efficiency and accuracy.

\subsection{Impacts of Real-World BIM Imperfections}
\label{sec:imperfections}
{The practical deployment of BIM-Loc must account for imperfections commonly present in real-world BIM models. A primary category is as-designed versus as-built discrepancies, where the constructed facility deviates from the design model. These represent the core challenge that BIM-Loc is specifically designed to address. By explicitly modeling which BIM surfaces are confirmed as built versus un-built through the discrepancy detection module, BIM-Loc treats these deviations as structured priors rather than sensor noise. This capability has been validated on the real-world CityU dataset, which contains real as-built conditions. These considerations demonstrate the practical boundaries within which BIM-Loc is expected to perform reliably.}

{Beyond as-built deviations, LOD also governs localization performance. BIM-Loc defaults to LOD~300 because this level provides sufficient geometric accuracy for primary structural elements and their topological relationships. Higher LODs ($>$~300) do not substantially improve structural geometric fidelity. After removing non-structural IFC entities, models above LOD~300 remain compatible. Lower LODs ($<$~300) degrade performance predictably. At LOD~200, dimensional tolerances and simplified geometry introduce misalignment into data association and trajectory optimization, which can be partially mitigated by relaxing the scan-to-BIM distance threshold at the cost of reduced accuracy. At LOD~100, the BIM models in city-level scenes lack requisite surface geometry and topological structure, rendering it unsuitable. We recommend at least LOD~300 for standard operation, with LOD~200 applicable via tuned parameters but with anticipated lower fidelity.}

\section{Limitations \& Future Works}
\label{sec:futurework}
While the proposed BIM-Loc system demonstrates reliable indoor localization performance by leveraging BIM's structural advantages and discrepancy detection capabilities, several limitations affect its broader applicability. First, the method requires an approximate initial position to initialize localization, restricting deployment in scenarios lacking prior positional information, such as initial exploration of unfamiliar buildings or emergency response missions. Second, the environment exploration process remains passive. {During human-operated inspections, the absence of real-time density visualization prevents operators from intuitively assessing whether a surface has been sufficiently scanned in large scale environments. This lack of immediate feedback often results in either incomplete coverage requiring rework or unnecessary redundant sampling, compromising both operational efficiency and the feasibility of full automation.}

{Furthermore, BIM-Loc is a domain-specific localization framework applied in BIM-available buildings. Meanwhile, we fully take advantages of BIM models in localization tasks instead of fine-level mapping. The design of discrepancy detection relates to the only presence or absence of building elements. The geometric modifications of existing elements, such as shape changes, boundary shifts, thickness variations, or partial deformations are not considered. BIM-Loc achieves higher precision and robustness in construction monitoring scenarios at the cost of dependency on BIM models.} {Although BIM-Loc localization capability is designed to be robust under missing/extra building structures with random shifts shown in Table~\ref{tab:ape_rmse_discrepancy}, BIM-Loc only provides structure-level existence checking instead of accurate mapping. It cannot provide quantitative analysis for amount of deviations. Instead, BIM-Loc focuses on robust localization under large real-BIM discrepancies.}

To address these limitations, future work will pursue three possible directions. First, we will develop a global localization method using BIM models alone, integrating topological structures with multi-sensor data (visual features, inertial measurements) to estimate absolute positions without initial guesses, expanding applicability to new environments. Second, we will create a BIM-adapted exploration algorithm that autonomously plans optimal inspection paths based on BIM geometry, dynamically adjusting via real-time coverage feedback to align algorithmic assessment with operational needs.These enhancements will significantly improve BIM-Loc's robustness and autonomy, better supporting applications in intelligent building maintenance, automated structural inspection, and autonomous patrols, advancing fully automated BIM-integrated structural health monitoring. {Finally, not only BIM models, the scope of prior-map based localization can be extent to more general representation, such as triangle meshes. Under this setting, the ray casting strategy can also be utilized and the limitations of application domains by reconstructed mesh prior maps.}

\section{Conclusions}
\label{sec:conclusion}
This paper presents BIM-Loc, a novel approach for accurate and robust localization in feature-sparse indoor environments where precise prior maps are unavailable. By using only as-designed BIM models, BIM-Loc overcomes the limitation of lacking pre-built maps through three key contributions: a multi-hit ray casting method for reliable online association between point clouds and BIM structures in 3D Cartesian and 2D texture spaces; a unified pose graph optimization framework integrating odometry, inter-scan consistency, and BIM-scan constraints to produce real-time BIM-aligned trajectories; and a Bayesian kernelized inference approach with hierarchical networks for incremental discrepancy identification from pixel to structural levels. Extensive evaluation on a custom BIM-robot simulator and real-world datasets shows that BIM-Loc significantly outperforms offline bundle adjustment methods and point-map-based approaches across five metrics. The results validate its robustness in sparse environments and effectiveness in providing both reliable localization and actionable discrepancy insights. BIM-Loc represents substantial progress toward practical BIM-integrated localization systems that operate throughout the building life cycle, from construction to completion, delivering both precise positioning and valuable structural monitoring capabilities.

\section*{Declaration of conflicting interests}
The authors declared no potential conflicts of interest with respect to the research, authorship, and/or publication of this article.

\section*{Funding}
This work was supported by the Natural Science Foundation of China (62461160309), the NSFC-RGC Joint Research Scheme (N\_HKU705/24), Hong Kong RGC (GRF 17201025, GRF17200924).

\bibliographystyle{SageH}

\begin{table}[t]
  \centering
  \caption{Symbol definition for BIM-Loc method.}
  \label{tab:symbol-description}
  \renewcommand{\arraystretch}{1.2}
  \small
  \begin{tabular*}{240pt}{@{\extracolsep\fill}p{2.0cm}>{\raggedright\arraybackslash}p{6.0cm}@{\extracolsep\fill}}
    \toprule[1.2pt] 
    \textbf{Symbol} & \textbf{Description} \\
    \midrule[0.8pt] 
    \multicolumn{2}{c}{\textit{General Formulation} (Section~\ref{sec:problem-formulation})} \\
    \midrule[0.5pt]
    $\boldsymbol{x}\coloneqq[\mathbf{R}\mid\boldsymbol{t}]$ & Sensor pose state, indexed by $k$. \\
    $\mathcal{X}\coloneqq\{\boldsymbol{x}_k\}_{k=1}^K$ & Trajectory with $K$ state sequences. \\
    $\mathcal{P}\coloneqq\{\boldsymbol{p}_i\}$ & Scan cloud points indexed by $i$. \\
    $\mathcal{Z}\coloneqq\{\boldsymbol{z}_k\}_{k=1}^K$ & Observation, $\boldsymbol{z}\coloneqq\{\boldsymbol{x},\mathcal{P}\}$ \\
    $\mathcal{B}\coloneqq\{\boldsymbol{b}_n\}_{n=1}^N$ & Structural components. \\
    $\mathcal{L}\coloneqq\{\mathbb{I}_n\}_{n=1}^N$ & Discrepancy indicator, $\mathbb{I}_n \in \{0, 1\}$. \\
    $\delta(n, k)$ & BIM-scan association indicator. \\
    \midrule[0.5pt]
    \multicolumn{2}{c}{\textit{Multi-Hit Ray Casting} (Section~\ref{sec:multi-hit})} \\
    \midrule[0.5pt]
    $\mathcal{R},\mathcal{H},\mathcal{M}$ & LiDAR ray set, hit set, miss set. \\
    $(u,v)$ & Barycentric coordinates. \\
    $\sigma\in\{0, 1\}$ & Intersection status: hit $\sigma=1$, miss $\sigma=0$. \\
    \midrule[0.5pt]
    \multicolumn{2}{c}{\textit{BIM-Aided Trajectory Optimization} (Section~\ref{sec:bim-factor})} \\
    \midrule[0.5pt]
    $\{\pi_f\}_{f=1}^F$ & Plane set for unique planar BIM facets. \\
    $\{\pi_c\}_{c=1}^C$ & Plane set for point clusters. \\
    $C_f$ & Number of point clusters associated to facet $f$, which follows the relationship $C=\sum_{f=1}^F{C_f}$. \\ 
    $\boldsymbol{\mu}_c, \mathbf{\Sigma}_c$ & Mean vector and covariance matrix of each point cluster. \\
    $\mathbf{Q}$ & Orthogonal matrix from the SVD decomposition of covariance matrix with respect to a point cluster. \\
    $\Lambda$ & Diagonal matrix for saving eigenvalues $\lambda_1, \lambda_2, \lambda_3$\\
    $\boldsymbol{e}_x, \boldsymbol{e}_y$ & Unit basis vector along $x$ and $y$ axis. \\
    $\boldsymbol{\mu}_g,\boldsymbol{n}_{g}, \mathbf{\Sigma}_g$ & Mean vector, normal vector, and covariance matrix from the aggregation of all associated point clusters. \\
    \midrule[0.5pt]
    \multicolumn{2}{c}{\textit{Discrepancy Detection} (Section~\ref{sec:discrepancy-identification})} \\
    \midrule[0.5pt]
    $\mathcal{F}, \mathcal{I}$ & Facet set and texture image set. \\ 
    $\mathbb{I}_\text{S}, \mathbb{I}_\text{F}, \mathbb{I}_\text{I}$ & Structure, facet, and pixel-wise discrepancy indicator. \\
    $\theta_\text{S}, \theta_{\text{F}}$ & Hyperparameters of $\mathbb{I}_\text{S}, \mathbb{I}_\text{F}$ distributions. \\
    $w_\text{F}\in[0, 1]$ & Area-based facet weight. \\ 
    $\alpha_\text{I}, \beta_\text{I}$ & Hyperparameters of pixel-wise discrepancy indicators. \\
    $k(\cdot, \cdot)$ & Kernel function. \\
    $\gamma^{+}_\mu, \gamma^{-}_\mu$ & Upper/lower bounds of estimated mean. \\
    $\gamma_{\sigma}$ & Threshold of estimated variance.  \\ 
    \bottomrule[1.2pt]
  \end{tabular*}
\end{table}

\section*{Appendix}

\appendix
\renewcommand{\appendixname}{Appendix~\Alph{section}}
\section{{Symbol Definition}}
The notations used in the BIM-Loc method are shown in Table~\ref{tab:symbol-description}. The indices for different parts of the method are defined as follows: index $k$ distinguishes pose states, observations, and scans. Index $n$ refers to structures in the BIM models. Index $i$  corresponds to a single point in each scan (i.e., $\boldsymbol{p}_i \in \mathcal{P}$). Index  $c$ identifies a point cluster; and index $f$ enumerates facets, primarily in Section~\ref{sec:bim-factor}.

\begin{figure}[t]
    \centering
    \includegraphics[width=1.0\linewidth]{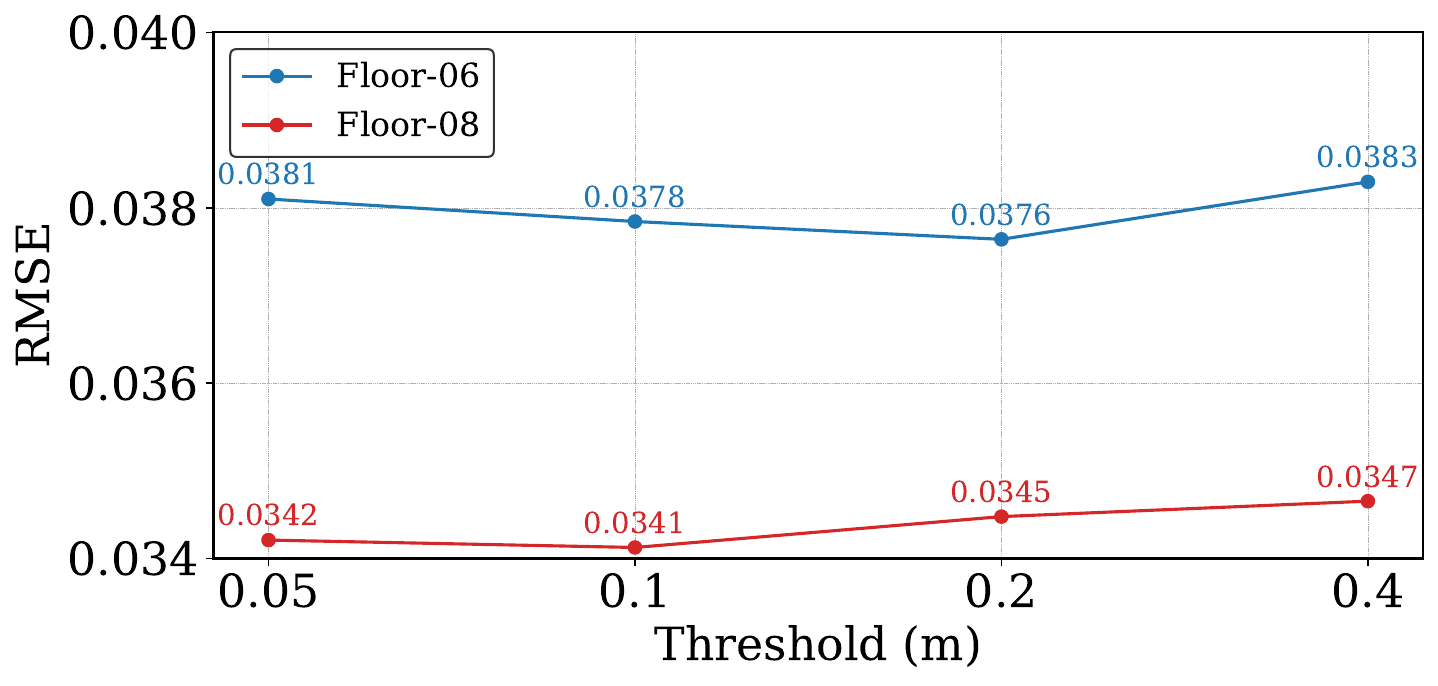}
    \caption{
    {Sensitivity analysis of scan-to-BIM distances on real-world CityU dataset.}
    }
    \label{fig:scan-to-bim-ablation}
\end{figure}

\begin{table}[t]
  \centering
  \caption{{Hyper-parameters sensitivity analysis in discrepancy detection. Parameters are given as ($\gamma_{\mu}^+$, $\gamma_{\mu}^-$, $\gamma_{\sigma}$).}}
  \label{tab:hyperparameters}
  \renewcommand{\arraystretch}{1.25}
  \small
  \begin{tabular}{@{}lcc@{}}
    \toprule[1.2pt]
    \textbf{Parameters} & \textbf{F1-Score} & \textbf{Scan-to-BIM RMSE (m)} \\
    \midrule[0.8pt]
    (0.55, 0.45, 16.0) & 0.848 & 0.0135 \\
    (0.65, 0.35, 8.0)  & 0.878 & 0.0132 \\
    (0.75, 0.25, 4.0)  & 0.893 & 0.0131 \\
    \underline{(0.85, 0.15, 2.0)}  & \textbf{0.913} & 0.0128 \\
    (0.95, 0.05, 1.0)  & 0.697 & \textbf{0.0127} \\
    \bottomrule[1.2pt]
  \end{tabular}
\end{table}

\section{Sensitivity Analysis}
\label{sec:sensitivity}
{Sensitivity analysis is conducted on real-world (CityU construction benchmark) and synthetic (CityU‑01) BIM models, considering noise in pose initialization. With these settings, the selected hyper‑parameters transfer seamlessly to other real‑world environments. Hyper‑parameter choices for BIM‑Loc are listed in Table~\ref{tab:hyperparameters} and Figure~\ref{fig:scan-to-bim-ablation}. As shown in Figure~\ref{fig:scan-to-bim-ablation}, RMSE follows a U‑shaped trend with the threshold, capturing the trade‑off between outlier rejection and geometric constraint sufficiency: a small threshold rejects true inliers under noisy initial poses, while a large threshold admits spurious point‑to‑BIM correspondences that degrade optimization. We select $0.20$~m as the default threshold to balance robustness against noisy initialization and effective outlier suppression. Table~\ref{tab:hyperparameters} shows that $(0.85, 0.15, 2.0)$ yields the best discrepancy‑detection performance (F1‑score $0.913$) with minimal impact on localization: the default scan‑to‑BIM RMSE of $0.0128$~m is only marginally above the minimum $0.0127$~m attained by the most restrictive setting $(0.95, 0.05, 1.0)$. Prioritizing F1‑score without significantly compromising localization accuracy, we adopt $(0.85, 0.15, 2.0)$ as the default parameter combination to achieve the best overall trade‑off.}

\section{Ablation Studies}
\label{sec:ablation}
\subsection{Discrepancy Detection Module}
\label{sec:ablation-des}
\begin{figure}[t]
    \centering
    \begin{subfigure}{\columnwidth}
        \centering
        \includegraphics[width=\linewidth]{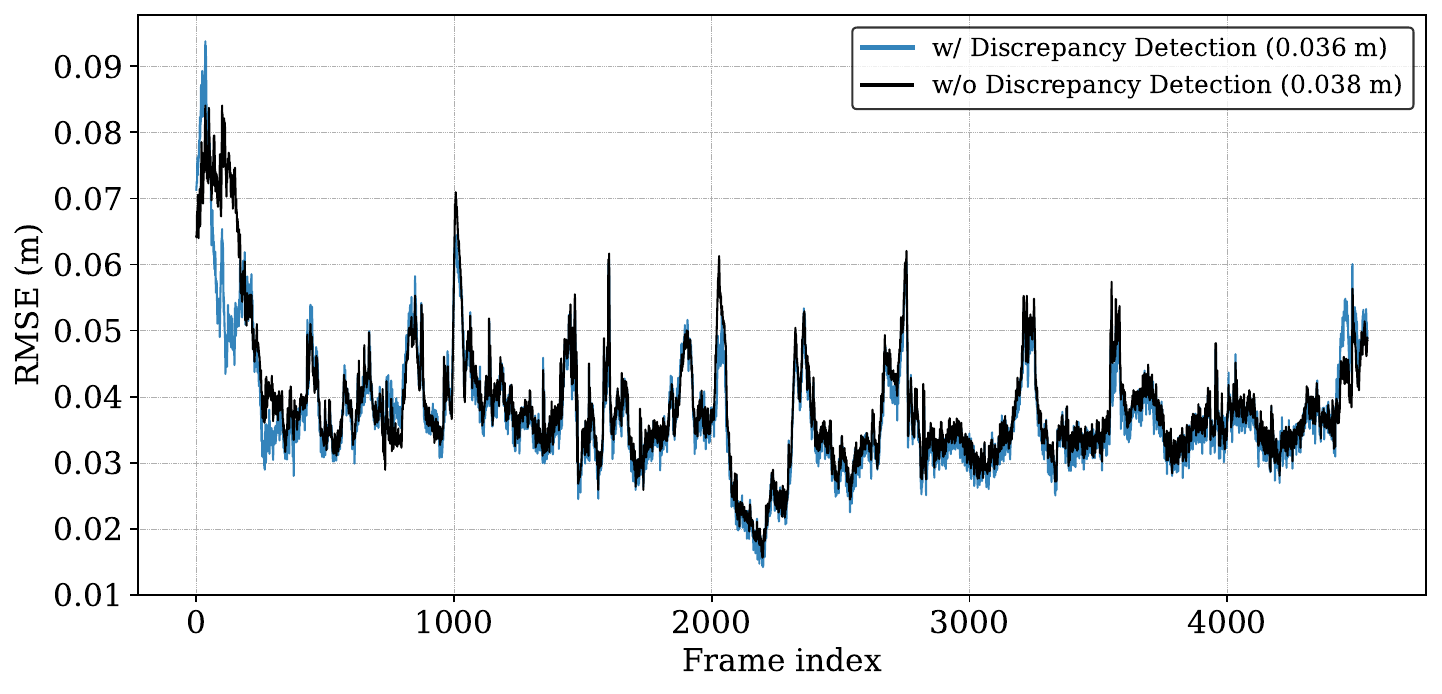}
        \caption{Floor 06}
    \end{subfigure}
    \begin{subfigure}{\columnwidth}
        \centering
        \includegraphics[width=\linewidth]{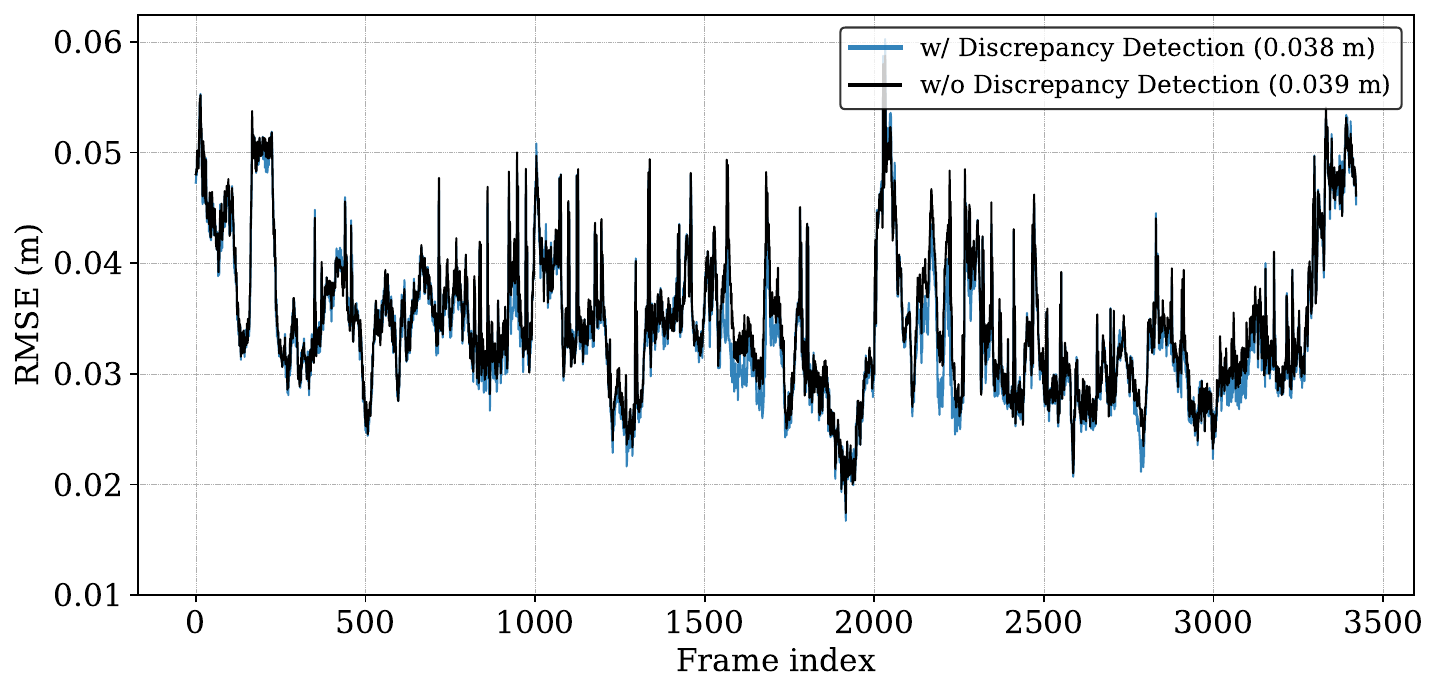}
        \caption{Floor 08}
    \end{subfigure}
    \caption{{Scan-to-BIM RMSE over time with and without the discrepancy detection module on CityU construction dataset with Livox Mid360 (Floor 06 and Floor 08).}}
    \label{fig:ablation_rmse}
\end{figure}

{To evaluate the functionality of discrepancy detection module, we compare BIM-Loc in two configurations: the BIM-Loc full system (with discrepancy detection) and a variant with the discrepancy module disabled. Quantitative results of Scan-to-BIM distance RMSE over time are given in Figure~\ref{fig:ablation_rmse} and Table~\ref{tab:mean_comparison}. On both Floor~06 and~08, the scan-to-BIM distance RMSE with the discrepancy detection module is slightly lower than that without it. On Floor~06, enabling the discrepancy module also accelerates convergence and keeps the error stable thereafter. This ablation helps isolate the benefit of the discrepancy module while demonstrating that the core localization pipeline remains effective even when operating on unfiltered scan-to-BIM data.}

{As show in Figure~\ref{fig:ablation_rmse}, the full BIM-Loc achieves slightly better RMSE than the version without the discrepancy module. Without removing non-built structures, the computational load is also higher than the full version, as the optimizer must process a larger set of correspondences including those from as-yet-unbuilt or inconsistent regions. As summarized in Table~\ref{tab:mean_comparison}, enabling discrepancy handling reduces the cost of multi-hit ray casting, filters out invalid factors, and improves the overall efficiency of the discrepancy detection during update.}

\begin{table}[t]
  \centering
  \caption{{Mean runtime duration of each module under with and without discrepancy (ms).}}
  \label{tab:mean_comparison}
  \renewcommand{\arraystretch}{1.25}
  \small
  \begin{tabular*}{240pt}{@{\extracolsep\fill}l*{2}{>{\centering\arraybackslash}p{2.3cm}}@{\extracolsep\fill}}
    \toprule[1.2pt]
    \textbf{Module} & \textbf{w/ Discrepancy} & \textbf{w/o Discrepancy} \\
    \midrule[0.8pt]
    Multi-Hit Ray Casting & 114.92 & 140.11 \\
    Factor Generation & 6.68 & 10.24 \\
    Trajectory Optimization & 20.85 & 29.87 \\
    Discrepancy Detection & 22.60 & 44.30 \\
    \bottomrule[1.2pt]
  \end{tabular*}
\end{table}

\begin{table}[t]
    \caption{
    {RMSE of ATE values for the simulation benchmark (spatial and temporal discrepancy tiers) in Floor 07, 08 in simulation benchmark. Each cell reports RMSE values for translation error [m] $\mid$ rotation error [degree]. Lower values indicate better performance.}
    }
    \label{tab:ape_rmse_discrepancy}
    \centering
    \renewcommand{\arraystretch}{1.25}
    \scriptsize
    \begin{tabular*}{240pt}{@{\extracolsep\fill}p{2.4cm}cc@{\extracolsep\fill}}
        \toprule[1.2pt]
        \footnotesize\textbf{Scene} & \footnotesize\textbf{BIM-Loc} & \footnotesize\textbf{DLO} \\
        \midrule
        Spatial Tier 1 (5\%)  & $\textbf{0.041}|\textbf{0.441}$ & $\underline{0.161}|\underline{1.103}$ \\
        Spatial Tier 2 (15\%) & $\textbf{0.028}|\textbf{0.510}$ & $\underline{0.159}|\underline{1.125}$ \\
        Spatial Tier 3 (25\%) & $\textbf{0.046}|\textbf{0.513}$ & $\underline{0.342}|\underline{1.159}$ \\
        \midrule
        Temporal Tier 1 (5\%)  & $\textbf{0.022}|\textbf{0.397}$ & $\underline{0.075}|\underline{0.586}$ \\
        Temporal Tier 2 (15\%) & $\textbf{0.016}|\textbf{0.184}$ & $\underline{0.057}|\underline{0.171}$ \\
        Temporal Tier 3 (35\%) & $\textbf{0.016}|\textbf{0.211}$ & $\underline{0.055}|\underline{0.155}$ \\
        \bottomrule[1.2pt]
    \end{tabular*}
\end{table}

\begin{figure}[t]
    \centering
    \includegraphics[width=\columnwidth]{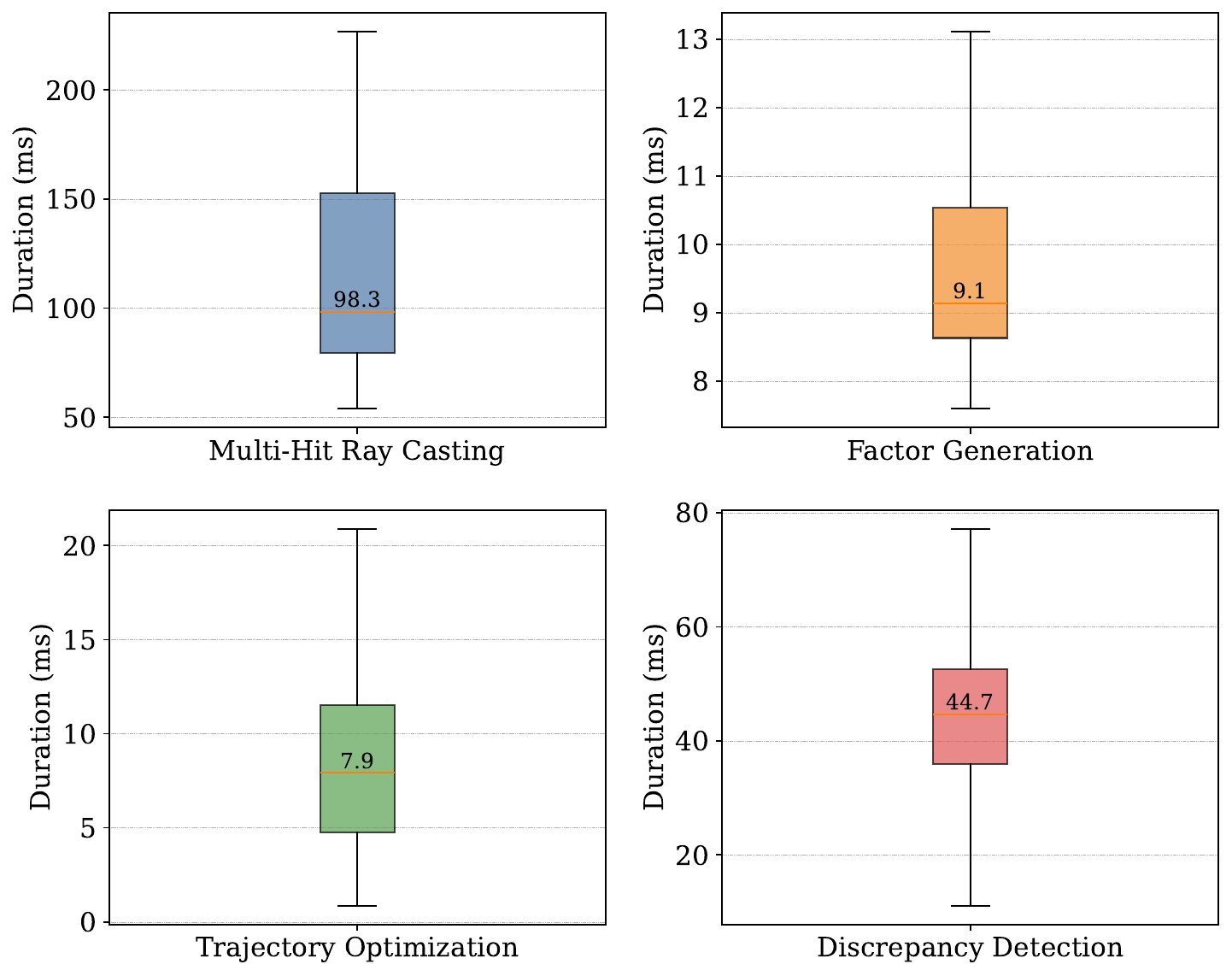}
    \caption{Runtime profiling of different pipeline sections, showing the distribution of per-scan execution time for each module.}
    \label{fig:runtime_profiler_sections_boxplot}
\end{figure}

\begin{figure*}[hbpt]
    \centering
    \begin{subfigure}[t]{0.49\linewidth}
        \centering
        \includegraphics[width=\linewidth]{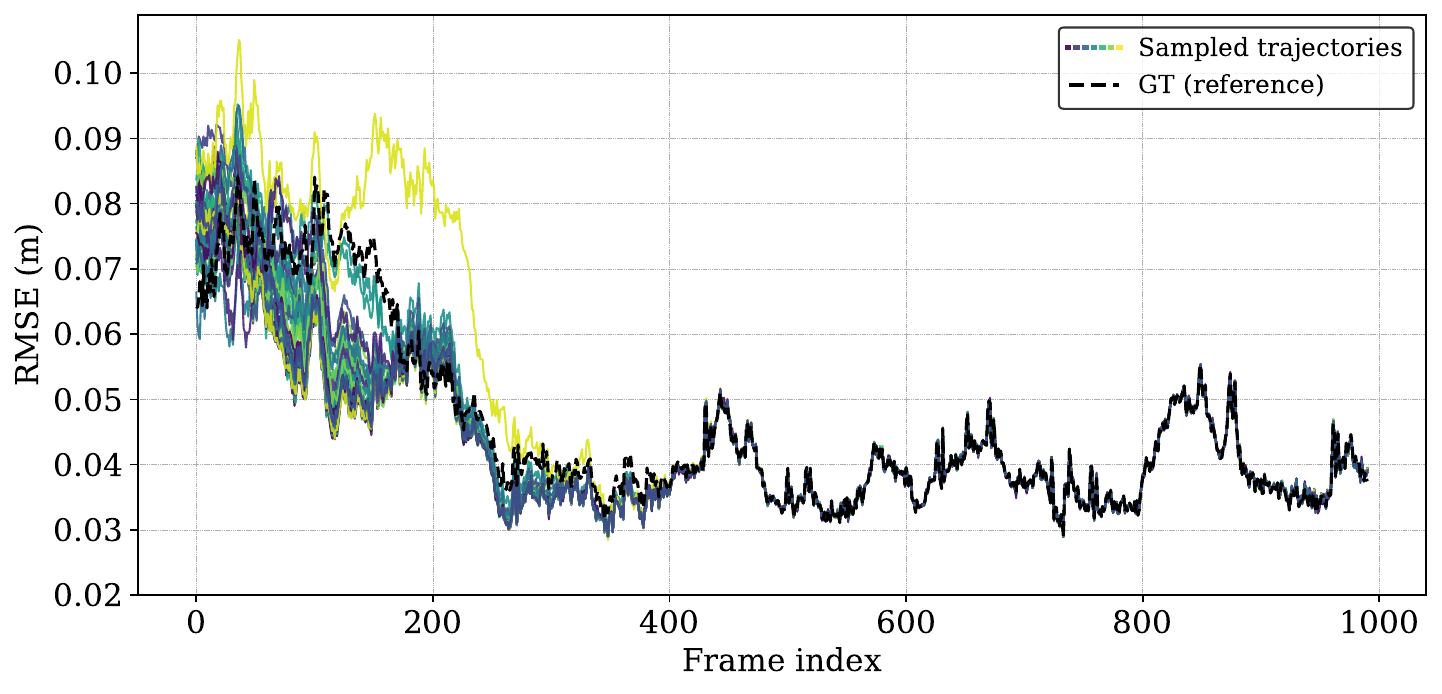}
        \caption{Floor~06: RMSE over time}
    \end{subfigure}\hfill
    \begin{subfigure}[t]{0.49\linewidth}
        \centering
        \includegraphics[width=\linewidth]{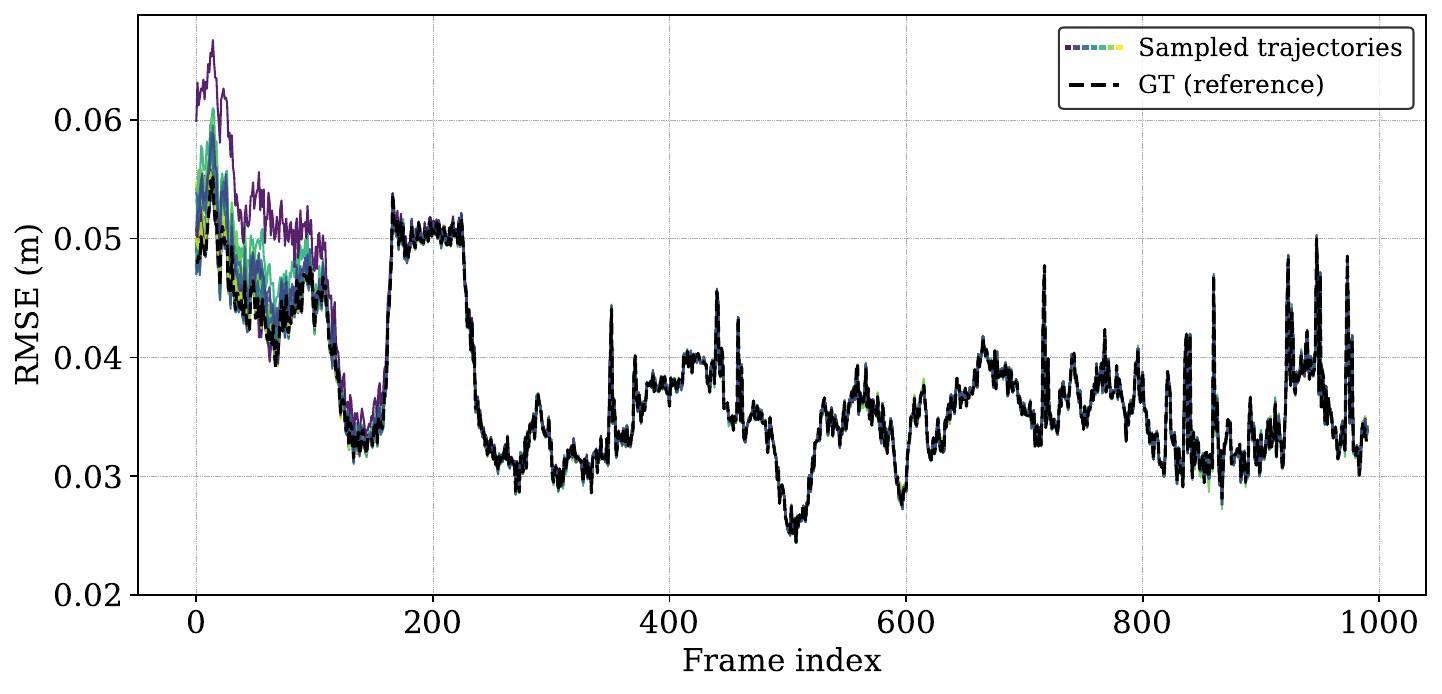}
        \caption{Floor~08: RMSE over time}
    \end{subfigure}

    \vspace{0.4em}
    \begin{subfigure}[t]{0.49\linewidth}
        \centering
        \includegraphics[width=\linewidth]{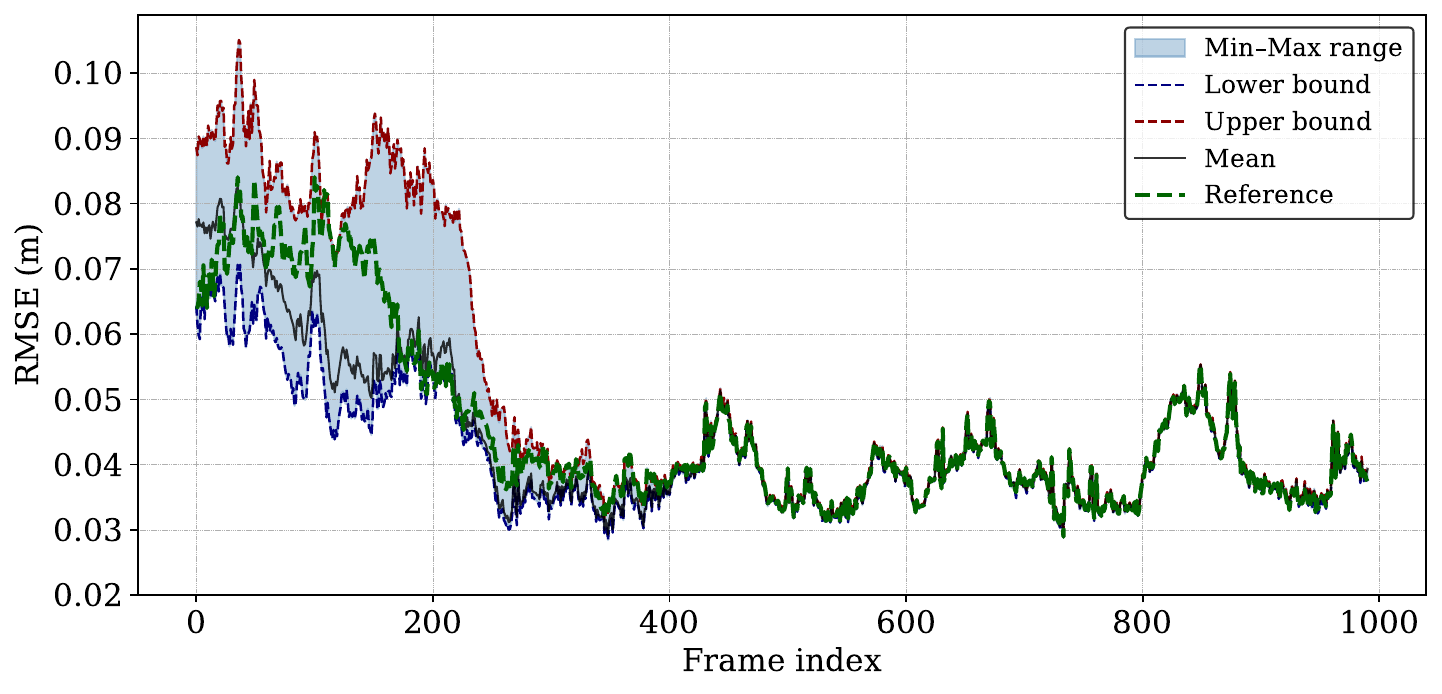}
        \caption{Floor~06: RMSE Distribution}
    \end{subfigure}\hfill
    \begin{subfigure}[t]{0.49\linewidth}
        \centering
        \includegraphics[width=\linewidth]{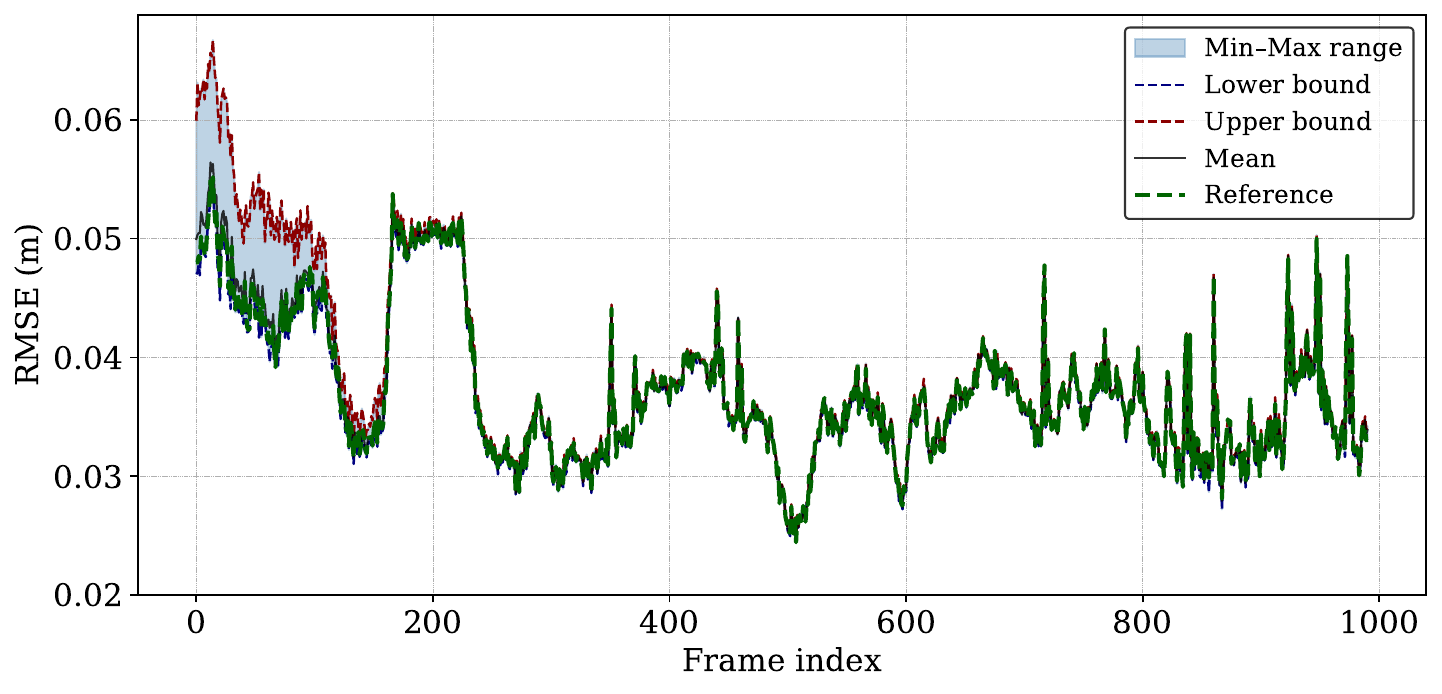}
        \caption{Floor~08: RMSE Distribution}
    \end{subfigure}

    \caption{{Initialization sensitivity on CityU dataset (Livox Mid360). (a)--(b) Scan-to-BIM distance RMSE for perturbed initializations on Floor~06 and Floor~08. (c)--(d) Range of RMSE across the 50 trials over time. The band narrows quickly, indicating low variance in final accuracy.}}
    \label{fig:initial-pose}
\end{figure*}

{Furthermore, to evaluate the performance boundary of BIM-Loc when dealing with BIM-real discrepancy, experiments under a series of controlled discrepancies are conducted like: (1) Missing structures: Up to \textbf{35\%} of the structural elements in as-designed BIM are missing. (2) Extra structures and Dimension errors: Up to \textbf{25\%} occlusion of views by extra objects. Besides, these building structures are randomly shifted in the range of [0, 0.5]~m. Table~\ref{tab:ape_rmse_discrepancy} reports the ATE RMSE of trajectories under different discrepancy levels, including three tiers with extra structures and random shifts (Addition tier) and three tiers with miss structures (Removal tier). According to experiment results. BIM-Loc maintains low error across all tiers and consistently outperforms the DLO baseline, supporting the above tolerance ranges.}

{Generally, reliable 6-DoF localization requires at least 3--4 non-coplanar structural facets (e.g., two walls and a floor). Below this (e.g., facing a single flat wall), the configuration becomes degenerate.}

\subsection{Initial Pose}
\label{sec:ablation-initial-pose}
{To assess the sensitivity of BIM-Loc to the quality of the initial pose, we conduct a controlled perturbation study on the CityU construction dataset using Livox Mid360 data. The nominal initial poses are obtained from a global scan-to-BIM registration method~\cite{Zhang2024AIC}, which typically achieves translation errors of 5--7~cm and rotational errors within 1$^\circ$ in experiments. Around this baseline for Floors~06 and~08, we sample 50 initial poses per floor from Gaussian distributions in $SE(3)$: (i) translation noise with $\mu = 0$ and $\sigma = 0.05$~m (so $3\sigma \approx 0.15$~m), and (ii) rotational noise with $\sigma = 3.33^\circ$ (so $3\sigma \approx 10^\circ$). Each perturbed initialization is used to run BIM-Loc for 100~s on the same sequence. As shown in Figure~\ref{fig:initial-pose}, the resulting scan-to-BIM distance RMSE curves rapidly converge and largely overlap, and the RMSE range across the 50 trials narrows within 10--30~s, indicating low variance in final accuracy and robustness to moderate initialization errors.}

\section{Runtime Analysis}
\label{sec:runtime-analysis}
{A runtime analysis is provided to validate the scalability of BIM-Loc. The runtime breakdown by pipeline sections (multi-hit ray casting, pose factor generation, trajectory optimization, discrepancy detection) is shown in Figure~\ref{fig:runtime_profiler_sections_boxplot}. BIM-Loc uses batch-wise processing: scan points are accumulated and the three batch-wise stages (multi-hit ray casting, pose factor generation, and discrepancy detection) are parallelized across threads, while trajectory optimization runs sequentially. With LiDAR at 10~Hz, the batch period is set to 1.5~s (15 frames). The profiler shows that per-frame cost stays within soft real-time bounds: batch processing takes about 350~ms per 1.5~s batch, and trajectory optimization about 22~ms per frame (100~ms cycle). Scalability with respect to BIM size, scan density, and environment scale is supported by the complexity arguments and by the fact that the evaluated sequences span multiple floors and BIM scales.}

\section{BIM Pre-Processing}
\label{sec:bim-preprocess}
{The BIM models used in the work are all with LOD~300 (i.e., Levels of Details) and come from the International Foundation Classes (IFC), which is a popular in AEC fields for efficient building data sharing. A series of tool chain is designed focusing on two functionalities: 1. transfer the raw IFC files into meshes of building elements, and 2. generate URDF files for usage in simulation platforms. In BIM/IFC files, the building structures are organized. Specifically, the raw IFC files are parsed automatically using the \textbf{IfcOpenShell} library, which systematically decomposes the architectural model into discrete building elements, including \texttt{IfcWall}, \texttt{IfcColumn}, \texttt{IfcBeam}, \texttt{IfcCovering}, \texttt{IfcSlab}, \texttt{IfcDoor}, \texttt{IfcWindow}, preserving both spatial hierarchies and semantic attributes. A rigorous filtering mechanism based on element type classification subsequently excludes non-structural entities such as MEP systems and \texttt{IfcFurniture}, ensuring only architecturally relevant load-bearing components participate in subsequent localization and mapping tasks. These selected geometric entities undergo robust polygon-to-triangle conversion (geometric processing module in IfcOpenShell) to generate computationally efficient watertight meshes, which are subsequently processed through ray casting and texture wrapping pipelines to produce visually consistent surface representations suitable for rendering. The resulting semantically-enriched mesh geometries, complete with their original hierarchical relationships and material attributes, are ultimately serialized into standardized URDF format to facilitate seamless integration into physics-based simulation environments, thereby enabling robust robot localization and automated discrepancy detection capabilities throughout the virtual building structure.}

\end{document}